%% file: iclr2027_conference.tex
\documentclass{article}
\usepackage{iclr2027_conference,times}

\input{math_commands.tex}

\usepackage{hyperref}
\usepackage{url}

\usepackage{bbm}
\usepackage{enumitem}
\usepackage{amssymb}
\usepackage{booktabs}
\usepackage{tabularx}
\usepackage[table]{xcolor}
\usepackage{array}
\usepackage{caption}
\usepackage{tcolorbox}
\usepackage{amsthm}
\usepackage{bm}
\usepackage{wrapfig}
\usepackage[utf8]{inputenc}
\usepackage{hyperref}
\usepackage{graphicx}
\usepackage{xcolor}
\usepackage{fontawesome5}

\usepackage{booktabs}
\usepackage{multirow}
\usepackage{graphicx}
\usepackage{adjustbox}
\usepackage{arydshln}
\usepackage{float}
\usepackage{etoc}

\definecolor{baselinebg}{RGB}{245,245,245}
\definecolor{oursbg}{RGB}{232,241,250}
\definecolor{sectioncolor}{RGB}{120,55,160}
\definecolor{poscolor}{RGB}{180,55,55}
\definecolor{negcolor}{RGB}{35,120,70}

\definecolor{projblue}{HTML}{2563EB}
\definecolor{softblack}{HTML}{333333}

\newcommand{\METHODNAME}{\textsc{SPACE}}

\title{Geometric Calibration of Supervised Fine-Tuning}

\author{
  Jianzhu Bao$^1$, 
  ~Jia-Chen Gu$^2$, 
  ~Qingyuan Liu$^1$,
  ~Haozhen Zhang$^1$ \\
  {\bf 
  ~Kai-Wei Chang$^2$,
  ~Nanyun Peng$^2$,
  ~Wenya Wang$^1$\thanks{\ Corresponding author.}
  } \\
  $^1$Nanyang Technological University ~
  $^2$University of California, Los Angeles \\
  \texttt{\{jianzhu.bao, qingyuan.liu, haozhen.zhang, wangwy\}@ntu.edu.sg} \\
  \texttt{\{gujc@, kwchang@cs., violetpeng@cs.\}ucla.edu} \\
  [3pt]
  {\normalsize
  \href{https://jianzhubao.github.io/SPACE/}
  {\textcolor{projblue}{\faGlobe}\hspace{2pt}Website}
  \quad
  \href{https://github.com/jianzhubao/SPACE}
  {\textcolor{softblack}{\faGithub}\hspace{2pt}Code}
  }
}

\iclrfinalcopy 
\begin{document}

\maketitle

\input{0-abstract}
\input{1-introduction}
\input{2-relatedWork}
\input{3-preliminaries}
\input{4-method}
\input{5-experiment}
\input{6-conclusion}
\input{7-statements}

\bibliography{iclr2027_conference}
\bibliographystyle{iclr2027_conference}

\clearpage
\appendix
\input{Appendix}

\end{document}

%% file: math_commands.tex
\usepackage{amsmath,amsfonts,bm}

\def\eqref#1{equation~\ref{#1}}

\def\1{\bm{1}}

\DeclareMathAlphabet{\mathsfit}{\encodingdefault}{\sfdefault}{m}{sl}
\SetMathAlphabet{\mathsfit}{bold}{\encodingdefault}{\sfdefault}{bx}{n}



%% file: 0-abstract.tex
\begin{abstract}
Recent studies investigate the weight geometry through its spectral structure, which characterizes how the weights transform different input directions.
Within this view, supervised fine-tuning (SFT) is shown to induce dense updates that perturb the spectral geometry of pre-SFT weights along their principal directions, potentially disrupting the model's prior knowledge.
These observations motivate using the geometry of pre-SFT weights as a reference for refining learned SFT updates.
We study this perspective as \mbox{\textit{Geometric Calibration}} and develop \METHODNAME{} (\textbf{S}ubspace \textbf{P}rojection for \textbf{A}dapted \textbf{C}heckpoint \textbf{E}nhancement), a lightweight post-hoc method that selectively reverts SFT update components within the principal subspace of the pre-SFT weights while preserving those in its orthogonal complement. 
\METHODNAME{} is a data-free, training-free method that requires only the pre- and post-SFT checkpoints.
We evaluate \METHODNAME{} on Qwen2.5, Qwen3, Qwen3.5, Llama-3.1, and Olmo-3 models ranging from 2B to 14B, across mathematical, coding, and medical SFT, as well as several SFT variants. Overall, \METHODNAME{} mitigates out-of-domain (OOD) capability regression while further improving in-domain (ID) performance over the uncalibrated SFT checkpoints. For math SFT, \METHODNAME{} yields average ID/OOD gains of 2.85/2.26 points on Qwen2.5-7B, and 2.41/1.52 points on Qwen3-14B-Base, respectively. These results highlight how the geometric relationship between SFT updates and pre-SFT weights shapes the adapted model's capabilities and, more broadly, motivate geometry-aware parameter updates for continual model improvement.
\end{abstract}

%% file: 1-introduction.tex
\begin{figure*}[!ht]
    \centering
    \includegraphics[width=\linewidth]{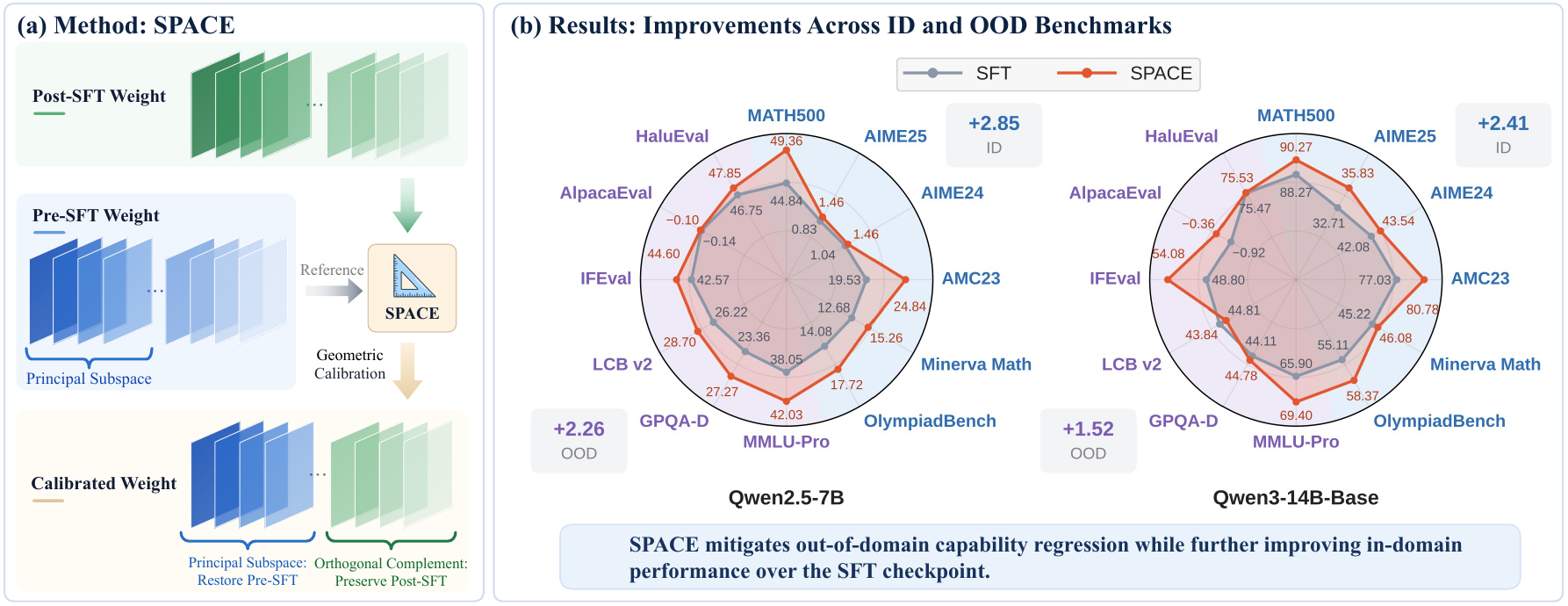}
    \caption{
\textbf{Overview of \METHODNAME{} and results.}
\textbf{(a) Method.} \METHODNAME{} uses the principal input subspace of the pre-SFT weight as a geometric reference for calibrating the post-SFT weights.
The pre-SFT weight is visualized as rank-one SVD components ordered by singular value, while the post-SFT weight is decomposed along the same pre-SFT input directions.
Blue and green denote pre- and post-SFT components, respectively, with color intensity reflecting the corresponding pre-SFT singular value.
\METHODNAME{} restores the pre-SFT transformation along the selected principal directions while retaining the post-SFT transformation along the remaining directions.
\textbf{(b) Results.} After regular SFT on mathematical data, \METHODNAME{} improves both in-domain (ID) and out-of-domain (OOD) performance over the uncalibrated SFT checkpoints.
    }
    \label{fig:teaser}
\end{figure*}

\section{Introduction}
\label{sec:introduction}

Supervised fine-tuning (SFT) is a standard approach for adapting large language models (LLMs) to target tasks, particularly for improving reasoning capabilities \citep{ho2023large,ye2025limo,lai2025survey}. However, gains on the target task can be accompanied by regression in capabilities beyond the training domain \citep{chu2025sft,huan2026does,lai2026reinforcement,shenfeld2026rl}, a phenomenon commonly studied as forgetting. To gain deeper insight into this problem, we believe it is important to understand how SFT changes the geometry of the model weights.

Recent work studies the weight geometry of LLMs through the spectral structure, which captures how different input directions are transformed \citep{zhu2025path,shen2026geometry,zhang2026geora}. Under this view, SFT tends to induce dense updates that perturb this geometry along principal directions. These principal directions typically correspond to singular vectors associated with large singular values, which together span a principal subspace. This subspace has been associated with important pre-existing or transferable capabilities, and perturbations within it can contribute to forgetting \citep{luo2026keeplora,wang2025milora,liu2026energy,jin2026rotation}.

These insights lead to a direct question: \emph{can an SFT checkpoint be improved by removing the learned updates within these principal subspaces?} More broadly, this motivates using pre-SFT weight geometry as a reference for refining completed SFT updates. We formalize this perspective as \mbox{\emph{Geometric Calibration}} and instantiate it with \METHODNAME{} (\textbf{S}ubspace \textbf{P}rojection for \textbf{A}dapted \textbf{C}heckpoint \textbf{E}nhancement), a lightweight post-hoc method for refining SFT checkpoints (Figure~\ref{fig:teaser}(a)). For each pre-SFT weight matrix, \METHODNAME{} defines a principal input subspace using its leading right singular vectors. It then projects the SFT update onto the orthogonal complement of this subspace, restoring the pre-SFT transformation along the principal directions while retaining the learned update along the remaining directions. Unlike existing post-hoc methods that determine refinement through parameter interpolation~\citep{wortsman2022robust,wang2025lines} or properties of the learned update itself~\citep{abdollahpoorrostam2026model,abro2026spectral}, \METHODNAME{} explicitly conditions the refinement on the geometric relationship between the SFT update and the pre-SFT weights. Moreover, \METHODNAME{} operates directly on the pre- and post-SFT checkpoints in a data- and training-free manner.

We evaluate \METHODNAME{} across five model families—Qwen2.5, Qwen3, Qwen3.5, Llama-3.1, and Olmo-3—ranging from 2B to 14B parameters, and across mathematical, coding, and medical SFT.
Across these settings, \METHODNAME{} consistently yields positive average gains over the corresponding SFT checkpoints on both ID and OOD evaluations.
For the two mathematical SFT settings in Figure~\ref{fig:teaser}(b), \METHODNAME{} improves average ID/OOD performance by 2.85/2.26 points on Qwen2.5-7B and 2.41/1.52 points on Qwen3-14B-Base.
The improvements further extend across different SFT variants, supporting the broad applicability of \METHODNAME{} beyond standard SFT.
It also outperforms a range of representative post-hoc model refinement baselines, highlighting the benefit of accounting for the relationship between SFT updates and pre-SFT geometry.
Overall, these results support the use of pre-SFT weight geometry as a reference for refining completed SFT updates and motivate the development of more targeted and effective geometric calibration methods for SFT.

%% file: 2-relatedWork.tex
\section{Related Work}
\label{sec:related_work}

\paragraph{Generalization and forgetting in SFT.}
Recent work has increasingly focused on understanding the generalization and forgetting behaviors of LLMs during SFT~\citep{wu2026generalization,huan2026does,ren2026rethinking}. Comparisons between SFT and reinforcement learning reveal distinct generalization and forgetting patterns~\citep{chu2025sft,shenfeld2026rl}. To mitigate capability degradation, existing efforts mainly improve SFT through changes to the training data or objectives. One line of work improves the selection and use of training data \citep{ye2025limo,kopiczko2026data}. Another line of work modifies the supervised objective through token reweighting or regularization~\citep{wu2026generalization,li2025beyond,zhu2026anchored,jin2026rotation}. These methods primarily intervene during SFT to control how the updates are learned. In contrast, we study whether a completed SFT update can be refined post hoc without data or additional training.

\paragraph{Weight geometry in model adaptation.}
Recent studies analyze the geometry of LLM weights during post-training through their spectral structure~\citep{zhu2025path,shen2026geometry,zhang2026geora}. 
These studies reveal that SFT typically produces dense updates along principal directions, which can distort the pretrained spectral structure. In contrast, reinforcement learning tends to induce sparse and off-principal updates, resulting in less disruption to the spectral structure~\citep{liu2025lift,mukherjee2026reinforcement}.
The role of principal directions has also been emphasized in studies of parameter-efficient fine-tuning (PEFT) and continual model merging.
In these lines of work, principal subspaces of pretrained weights are often associated with prior or transferable capabilities.
Accordingly, many PEFT methods organize adaptation around principal or complementary singular subspaces~\citep{meng2024pissa,wang2025milora,luo2026keeplora,xiong2026oplora}, while continual model merging methods leverage structured subspaces to reduce interference with previously acquired capabilities~\citep{ilharco2023editing,yu2024language,sun2026revisiting,li2026model,shi2026pact,qiu2026null}.
Together, these studies suggest that the geometry of pretrained weights provides a useful reference for controlling how new updates interact with existing capabilities.
Our work builds on this perspective but uses the principal subspace of the pre-SFT weights to refine a single completed SFT update post hoc.

\paragraph{Post-hoc model refinement.}
Our work is closely related to post-hoc model merging/editing/refinement methods that improve a trained model by directly modifying its parameters~\citep{wortsman2022robust,bandarkar2025layer,huang2025mitigating,bandarkar2025unreasonable}.
Existing approaches refine completed checkpoints through weight interpolation~\citep{wortsman2022robust,wang2025lines}, update pruning~\citep{yu2024language,huang2025mitigating}, model souping~\citep{wortsman2022model,abdollahpoorrostam2026model}, or spectral filtering~\citep{abro2026spectral}.
Their refinement strategies are typically determined by parameter interpolation or by properties of the learned update or the fine-tuned weights themselves.
Although some methods use the pre-SFT checkpoint, they generally do not explicitly consider its geometric relationship with the learned update when determining which update components should be modified.
Motivated by the importance of pre-SFT weight geometry discussed above, we instead study this relationship directly and use it to refine completed SFT checkpoints post hoc.

%% file: 3-preliminaries.tex
\section{Problem Formulation}
\label{sec:problem_formulation}

\paragraph{Supervised fine-tuning.} SFT adapts an LLM to target tasks by learning from labeled training data. Prior work has shown that SFT can induce non-trivial parameter updates \citep{yuan2025mitigating,mukherjee2026reinforcement}. While such updates can improve performance on the target task, they may also lead to capability regression beyond the training domain \citep{wu2026mitigating,shenfeld2026rl,huan2026does}. These observations suggest that some components of the learned SFT updates may be redundant or detrimental to the adapted model. We therefore investigate whether identifying and calibrating such components can mitigate capability regression while preserving or further improving target-domain performance.

\paragraph{Geometric calibration.}

Given a pre-SFT checkpoint \(\theta_0\) and its corresponding post-SFT checkpoint \(\theta_{\mathrm{SFT}}\), the SFT-induced parameter update is defined as
\begin{equation}
\Delta\theta_{\mathrm{SFT}}
=
\theta_{\mathrm{SFT}}-\theta_0.
\label{eq:sft_update}
\end{equation}

Rather than treating the post-SFT checkpoint as the endpoint of adaptation, we explore further refinement of the learned update to form the final adapted model. 
We approach this refinement through \emph{Geometric Calibration}, using the geometric structure of the pre-SFT weights as a reference.

Formally, we define a calibration operator \(\mathcal{C}\) that maps the learned SFT update to a calibrated update conditioned on the pre-SFT parameters:
\begin{equation}
\Delta\theta_{\mathrm{cal}}
=
\mathcal{C}
\left(
\Delta\theta_{\mathrm{SFT}};\theta_0
\right),
\label{eq:calibration_operator}
\end{equation}
where $\Delta\theta_{\mathrm{SFT}}$ is the update being calibrated and \(\theta_0\) serves as the geometric reference.
We seek a calibrated checkpoint \(\theta_{\mathrm{cal}}=\theta_0+\Delta\theta_{\mathrm{cal}}\) with improved overall capabilities.

%% file: 4-method.tex
\section{Method}
\label{sec:method}

We develop \METHODNAME{} as a post-hoc implementation for Geometric Calibration. For a pre-SFT weight
\(W_0\in\mathbb{R}^{d_{\mathrm{out}}\times d_{\mathrm{in}}}\)
and its post-SFT counterpart \(W_{\mathrm{SFT}}\), let \(\Delta W = W_{\mathrm{SFT}}-W_0\). The central idea is to refine the completed update \(\Delta W\) using the principal subspace of \(W_0\) as a geometric reference.

\subsection{Motivation}
\label{subsec:motivation}

\paragraph{Principal directions and capability preservation.} A common view in existing work is to characterize the geometry of model weights through their singular directions, where directions associated with large singular values are referred to as principal directions and their span defines the principal subspace~\citep{zhu2025path,wang2025milora,shen2026geometry}. Prior work suggests that this principal subspace is closely associated with important prior or broadly transferable knowledge, and that limiting changes to these directions can help preserve existing capabilities during adaptation \citep{wang2025milora,luo2026keeplora,sun2026revisiting,jin2026rotation}.
At the same time, SFT tends to induce dense updates that perturb the pre-SFT weight geometry within the principal subspace~\citep{zhu2025path,shen2026geometry}. Together, these findings motivate using the principal subspace of the pre-SFT weights as a reference for refining the learned SFT update. In particular, a natural calibration objective is to preserve the pre-SFT transformation within this subspace while retaining the post-SFT transformation along the remaining directions.

\paragraph{Preserving the pre-SFT transformation.}
For a layer input representation \(h\in\mathbb{R}^{d_{\mathrm{in}}}\), we seek a calibrated transformation \(W_{\mathrm{cal}}\) that follows the pre-SFT transformation on the component associated with the principal subspace of \(W_0\), while retaining the post-SFT transformation on the remaining component. Let \(P_0\) be a rank-\(k\) orthogonal projector onto the principal subspace, where \(k\) denotes its dimension. The desired behavior is
\begin{equation}
    W_{\mathrm{cal}}h
    =
    W_0P_0h
    +
    W_{\mathrm{SFT}}(I-P_0)h.
    \label{eq:calibration_behavior}
\end{equation}

To determine \(P_0\), we seek a projection that retains as much of the pre-SFT transformation as possible.
Specifically, among all rank-\(k\) orthogonal projectors, we choose \(P_0\) such that \(W_0P_0\) best approximates \(W_0\) in Frobenius norm:
\begin{equation}
    P_0
    \in
    \underset{\substack{
        P=P^\top=P^2\\
        \operatorname{rank}(P)=k
    }}{\arg\min}
    \left\|W_0-W_0P\right\|_F^2.
    \label{eq:principal_reference}
\end{equation}
Since \(P\) is an orthogonal projector, minimizing this error is
equivalent to maximizing
\(\operatorname{tr}(PW_0^\top W_0)\).
Let $W_0=U_0\Sigma_0V_0^\top$ be the SVD of $W_0$. The Ky Fan maximum principle
\citep{fan1949theorem} gives a solution
\begin{equation}
    P_0=V_{0,k}V_{0,k}^\top,
    \label{eq:pre_sft_projector}
\end{equation}
where \(V_{0,k}\) contains the leading \(k\) right singular vectors of
\(W_0\).
A complete derivation from \eqref{eq:principal_reference} to \eqref{eq:pre_sft_projector} is provided in Appendix~\ref{app:principal_input_projector}.
With this choice, \(W_0P_0=U_{0,k}\Sigma_{0,k}V_{0,k}^\top\) is a best rank-\(k\) approximation of \(W_0\). Thus, \(P_0h\) selects the component of the input representation associated with this principal part of the pre-SFT transformation. We refer to \(\operatorname{span}(V_{0,k})\) as the \emph{principal input subspace}, and use \(P_0\) for calibration.

\subsection{\METHODNAME{}: Subspace Projection for Adapted Checkpoint Enhancement}
\label{subsec:space}

\paragraph{Subspace Projection.}
Requiring~\eqref{eq:calibration_behavior} to hold for any layer input representation \(h\) gives \(W_{\mathrm{cal}}=W_0P_0+W_{\mathrm{SFT}}(I-P_0)\).
Using \(W_{\mathrm{SFT}}=W_0+\Delta W\), this can be equivalently written as
\begin{equation}
    W_{\mathrm{cal}}
    =
    W_0+\Delta W(I-P_0).
    \label{eq:calibrated_weight}
\end{equation}

Applying this calibration to the selected weight matrices yields the model-level calibrated update \(\Delta\theta_{\mathrm{cal}}\) defined in Section~\ref{sec:problem_formulation}. \METHODNAME{} therefore removes the SFT update within the principal input subspace while preserving it on the orthogonal complement.

\paragraph{Adaptive subspace dimension.} Since weight matrices in LLMs can differ substantially in both their dimensions and spectral concentration, using a fixed \(k\) across all matrices may capture very different fractions of their singular-value spectra. We therefore choose the subspace dimension \(k\) adaptively according to the singular-value spectrum of each weight matrix. 
Specifically, let \(\sigma_{0,1}\geq\cdots\geq\sigma_{0,r}\geq0\) denote the singular values of \(W_0\). We select the smallest \(k\) such that the leading singular components account for at least a fraction \(\rho\) of the total spectral energy:
\begin{equation}
    k
    =
    \min\left\{
        m:
        \frac{\sum_{i=1}^{m}\sigma_{0,i}^2}
             {\sum_{i=1}^{r}\sigma_{0,i}^2}
        \geq \rho
    \right\}.
    \label{eq:subspace_dimension}
\end{equation}

This allows the selected subspace to capture a comparable fraction of the dominant singular structure across different weight matrices. In practice, we use \(\rho=0.5\) and apply the projection to all two-dimensional weight matrices by default, as these choices work well across the evaluated settings.

%% file: 5-experiment.tex
\section{Experiments}
\subsection{Main Experiments on Mathematical Reasoning SFT}
\label{main_experiments_on_mathematical_reasoning_sft}

We first evaluate \METHODNAME{} in the setting of mathematical reasoning SFT, in which base models are fine-tuned on long chain-of-thought (CoT) mathematical reasoning data.

\begin{table*}[t]
\centering
\caption{
Main experimental results on mathematical reasoning SFT.
We compare ID and OOD performance across different post-hoc methods.
``Avg. (\(\Delta\) vs. SFT)'' denotes the average score change relative to the SFT checkpoint. \textbf{Bold} denotes the best score excluding the Base model.
}
\label{tab:main_results}

\scriptsize
\setlength{\tabcolsep}{3.2pt}

\begin{tabular}{
cl
>{\color{gray}}c
c
>{\columncolor{baselinebg}}c
>{\columncolor{baselinebg}}c
>{\columncolor{baselinebg}}c
>{\columncolor{baselinebg}}c
>{\columncolor{baselinebg}}c
>{\columncolor{baselinebg}}c
>{\columncolor{baselinebg}}c
>{\columncolor{baselinebg}}c
>{\columncolor{oursbg}}c
}
\toprule

\textbf{Model}
& \textbf{Benchmark}
& \textbf{Base}
& \textbf{SFT}
& \textbf{WiSE-FT}
& \textbf{TIES}
& \textbf{LiNeS}
& \textbf{DARE}
& \textbf{POME}
& \textbf{FAPM}
& \textbf{DG-Hard}
& \textbf{MonoSoup}
& \textbf{\METHODNAME{}}
\\

\midrule

\multirow{17}{*}{\rotatebox{90}{\textbf{Qwen2.5-7B}}}
& \multicolumn{10}{l}{\textcolor{sectioncolor}{\textit{In-domain (ID)}}} \\

& MATH500
& 28.79 & 44.84 & 45.14 & 45.70 & 43.89 & 44.13 & 46.23 & 46.83 & 47.50 & 46.43 & \textbf{49.36} \\

& AIME25
& 0.42 & 0.83 & 0.83 & 0.42 & 0.63 & 0.42 & 0.83 & \textbf{1.46} & 0.63 & 1.04 & \textbf{1.46} \\

& AIME24
& 1.65 & 1.04 & 1.04 & 0.63 & 0.83 & 0.83 & 1.04 & \textbf{1.67} & 1.46 & 1.46 & 1.46 \\

& OlympiadBench
& 7.69 & 14.08 & 17.06 & 14.62 & 14.34 & 13.82 & 15.75 & 17.44 & 15.39 & 15.06 & \textbf{17.72} \\

& Minerva Math
& 9.26 & 12.68 & 11.60 & 12.82 & 10.62 & 11.53 & 14.27 & 11.79 & 13.26 & 12.68 & \textbf{15.26} \\

& AMC23
& 15.65 & 19.53 & 22.03 & 19.38 & 20.00 & 18.28 & 20.94 & \textbf{24.84} & 22.97 & 20.00 & \textbf{24.84} \\

& \textbf{Avg. (\(\Delta\) vs. SFT)}
& -
& -
& \textcolor{poscolor}{+0.78}
& \textcolor{poscolor}{+0.10}
& \textcolor{negcolor}{-0.45}
& \textcolor{negcolor}{-0.67}
& \textcolor{poscolor}{+1.01}
& \textcolor{poscolor}{+1.84}
& \textcolor{poscolor}{+1.37}
& \textcolor{poscolor}{+0.61}
& \textbf{\textcolor{poscolor}{+2.85}}
\\

\cmidrule(lr){2-13}

& \multicolumn{10}{l}{\textcolor{sectioncolor}{\textit{Out-of-domain (OOD)}}} \\

& MMLU-Pro
& 40.33 & 38.05 & 40.85 & 38.12 & 41.90 & 38.11 & 39.41 & 38.36 & 35.16 & 39.99 & \textbf{42.03} \\

& GPQA-D
& 26.77 & 23.36 & 26.77 & 23.96 & 26.42 & 22.73 & 24.37 & 25.95 & 24.34 & 25.44 & \textbf{27.27} \\

& LCB v2
& 28.96 & 26.22 & \textbf{29.03} & 26.74 & 28.83 & 26.48 & 27.92 & 26.61 & 28.64 & 28.25 & 28.70 \\

& IFEval
& 42.57 & 42.57 & 45.56 & 42.57 & 43.88 & 41.25 & 42.81 & \textbf{45.68} & 41.85 & 45.44 & 44.60 \\

& AlpacaEval
& -1.13 & -0.14 & -0.20 & -0.13 & -0.12 & -0.11 & -0.15 & -0.16 & -0.20 & -0.16 & \textbf{-0.10} \\

& HaluEval
& 46.86 & 46.75 & 47.80 & 46.71 & 47.67 & 46.56 & 46.36 & 46.49 & 47.15 & 47.23 & \textbf{47.85} \\

& \textbf{Avg. (\(\Delta\) vs. SFT)}
& -
& -
& \textcolor{poscolor}{+2.17}
& \textcolor{poscolor}{+0.19}
& \textcolor{poscolor}{+1.96}
& \textcolor{negcolor}{-0.30}
& \textcolor{poscolor}{+0.65}
& \textcolor{poscolor}{+1.02}
& \textcolor{poscolor}{+0.02}
& \textcolor{poscolor}{+1.56}
& \textbf{\textcolor{poscolor}{+2.26}}
\\

\midrule

\multirow{17}{*}{\rotatebox{90}{\textbf{Qwen3-14B-Base}}}
& \multicolumn{10}{l}{\textcolor{sectioncolor}{\textit{In-domain (ID)}}} \\

& MATH500
& 55.07 & 88.27 & 88.60 & 88.60 & 89.47 & 88.20 & 89.40 & 88.20 & 88.93 & 89.20 & \textbf{90.27} \\

& AIME25
& 5.21 & 32.71 & 29.58 & 36.25 & 33.96 & 34.79 & 34.58 & 31.25 & 33.33 & \textbf{36.67} & 35.83 \\

& AIME24
& 6.25 & 42.08 & 38.54 & 41.46 & \textbf{45.21} & 42.92 & 44.38 & 35.83 & 42.92 & 41.88 & 43.54 \\

& OlympiadBench
& 27.31 & 55.11 & 55.85 & 56.74 & 57.33 & 55.36 & 57.14 & 55.56 & 54.72 & 56.25 & \textbf{58.37} \\

& Minerva Math
& 22.92 & 45.22 & 42.40 & 45.71 & 43.87 & 45.71 & 44.24 & 42.52 & 44.98 & 44.49 & \textbf{46.08} \\

& AMC23
& 38.59 & 77.03 & 78.59 & 79.84 & \textbf{80.78} & 80.31 & 79.38 & 78.28 & 80.00 & 80.16 & \textbf{80.78} \\

& \textbf{Avg. (\(\Delta\) vs. SFT)}
& -
& -
& \textcolor{negcolor}{-1.14}
& \textcolor{poscolor}{+1.36}
& \textcolor{poscolor}{+1.70}
& \textcolor{poscolor}{+1.15}
& \textcolor{poscolor}{+1.45}
& \textcolor{negcolor}{-1.46}
& \textcolor{poscolor}{+0.74}
& \textcolor{poscolor}{+1.37}
& \textbf{\textcolor{poscolor}{+2.41}}
\\

\cmidrule(lr){2-13}

& \multicolumn{10}{l}{\textcolor{sectioncolor}{\textit{Out-of-domain (OOD)}}} \\

& MMLU-Pro
& 61.80 & 65.90 & 66.30 & 69.20 & 66.20 & 67.00 & 68.10 & 67.00 & \textbf{69.40} & 67.80 & \textbf{69.40} \\

& GPQA-D
& 44.61 & 44.11 & 45.29 & \textbf{45.62} & 44.28 & 42.93 & 40.24 & 43.94 & 42.59 & 44.95 & 44.78 \\

& LCB v2
& 46.71 & 44.81 & 44.88 & 44.10 & 44.55 & 45.08 & \textbf{45.47} & 45.01 & 43.77 & 44.81 & 43.84 \\

& IFEval
& 65.35 & 48.80 & 48.20 & 51.80 & 47.48 & 52.16 & 52.04 & 45.08 & 50.00 & 51.20 & \textbf{54.08} \\

& AlpacaEval
& 0.42 & -0.92 & \textbf{-0.18} & -0.50 & -0.95 & -1.01 & -0.78 & -0.27 & -0.79 & -0.55 & -0.36 \\

& HaluEval
& 53.97 & 75.47 & 73.10 & 75.53 & 75.70 & 75.80 & \textbf{76.87} & 72.07 & 75.93 & 75.83 & 75.53 \\

& \textbf{Avg. (\(\Delta\) vs. SFT)}
& -
& -
& \textcolor{negcolor}{-0.10}
& \textcolor{poscolor}{+1.26}
& \textcolor{negcolor}{-0.15}
& \textcolor{poscolor}{+0.63}
& \textcolor{poscolor}{+0.63}
& \textcolor{negcolor}{-0.89}
& \textcolor{poscolor}{+0.46}
& \textcolor{poscolor}{+0.98}
& \textbf{\textcolor{poscolor}{+1.52}}
\\

\bottomrule
\end{tabular}
\end{table*}

\paragraph{Training recipe.}
We conduct full-parameter SFT on two base models: Qwen2.5-7B \citep{qwen2025qwen25technicalreport} and Qwen3-14B-Base \citep{yang2025qwen3}. Following prior studies \citep{zhu2026anchored,ren2026rethinking}, we use pretrained checkpoints without instruction tuning. For Qwen2.5-7B, following \citet{wu2026generalization,zhu2026anchored}, we use 100K training examples sampled from NuminaMath-CoT \citep{li2024numinamath}. For Qwen3-14B-Base, we follow \citet{ren2026rethinking} and use Math-CoT-20k, which contains 20K examples. Additional training details are provided in Appendix~\ref{app:detailed_training_settings_of_the_main_experiments}.

\paragraph{Benchmarks and evaluation protocol.}
We evaluate ID performance on six mathematical benchmarks: MATH-500 \citep{hendrycks2021measuring}, Minerva Math \citep{lewkowycz2022solving}, OlympiadBench \citep{he2024olympiadbench}, AIME 2024 \citep{aime24}, AIME 2025 \citep{aime25}, and AMC 2023 \citep{maa2023amc}. Following the benchmark suite used by \citet{ren2026rethinking}, we further evaluate OOD performance on six benchmarks: MMLU-Pro \citep{wang2024mmlu}, GPQA-Diamond (GPQA-D) \citep{rein2024gpqa}, LiveCodeBench v2 (LCB v2) \citep{jain2025livecodebench}, IFEval \citep{zhou2023instruction}, AlpacaEval (RM) \citep{dubois2024length}, and HaluEval \citep{li2023halueval}. Further evaluation details are provided in Appendix~\ref{app:detailed_evaluation_settings_of_the_main_experiments}.

\paragraph{Baselines.}
We compare \METHODNAME{} with eight representative post-hoc model editing/merging baselines:
WiSE-FT~\citep{wortsman2022robust},
TIES~\citep{yadav2023ties},
LiNeS~\citep{wang2025lines},
DARE~\citep{yu2024language},
POME~\citep{liu2025pome},
FAPM~\citep{huang2025mitigating},
DG-Hard~\citep{abro2026spectral},
and MonoSoup~\citep{abdollahpoorrostam2026model}.
See Appendix~\ref{app:baseline_details} for details.

\paragraph{Main results.}
Table~\ref{tab:main_results} shows that \METHODNAME{} yields the strongest overall improvements among all baselines, achieving the highest ``Avg. (\(\Delta\) vs. SFT)'' in both ID and OOD performance for both models.
On Qwen2.5-7B, \METHODNAME{} improves every evaluated benchmark over SFT and even surpasses the Base model on five of the six OOD benchmarks, suggesting that reasoning capabilities acquired during mathematical SFT may generalize to OOD tasks after calibration.
On Qwen3-14B-Base, \METHODNAME{} also yields substantial additional gains despite the already strong mathematical reasoning performance after SFT.
Importantly, the ID gains suggest that \METHODNAME{} does more than recover capabilities degraded by SFT. They also suggest that reverting updates within the principal subspace can improve generalization to unseen ID examples.
The comparison with WiSE-FT further highlights the benefit of selective calibration: global interpolation can recover OOD performance in some settings, but often provides smaller ID gains or degrades the adapted model, whereas \METHODNAME{} improves both.
Finally, \METHODNAME{}'s advantage over spectral post-hoc baselines such as DG-Hard and MonoSoup, which operate on the learned update itself, suggests that the geometry of the pre-SFT weights provides additional information for post-hoc refinement.
Overall, these results suggest that effective post-hoc refinement depends not simply on shrinking or restructuring the learned update, but on strategically calibrating it with respect to the geometry of the pre-SFT weights.

\begin{figure*}[t]
    \centering
    \includegraphics[width=\linewidth]{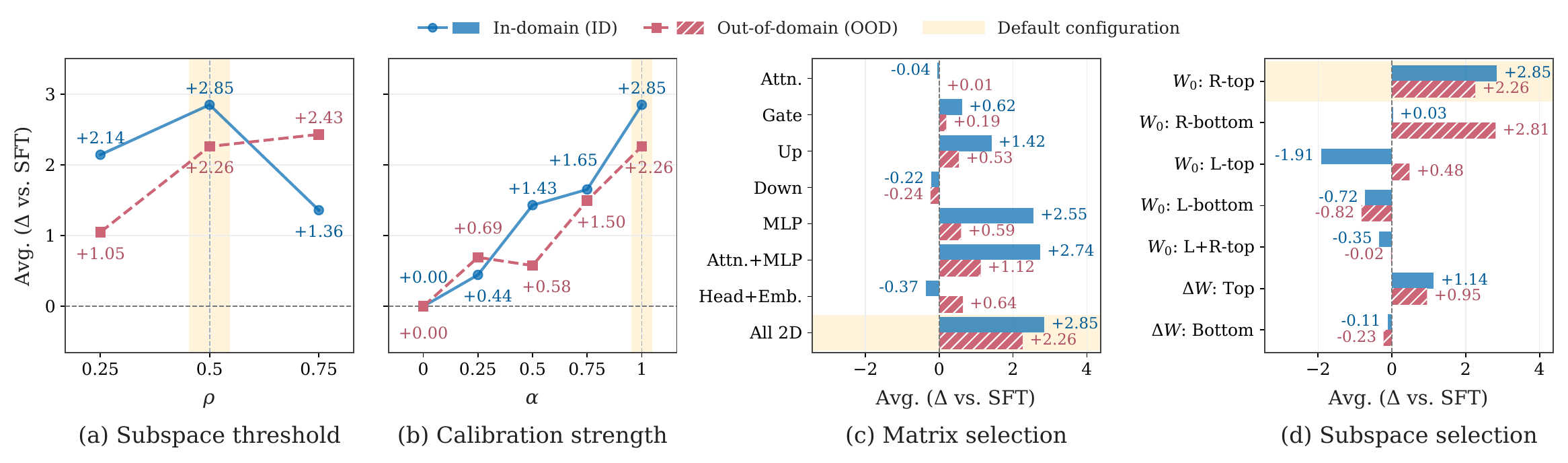}
\caption{
Ablation study on Qwen2.5-7B.
``Avg. (\(\Delta\) vs. SFT)'' denotes the average changes relative to the SFT checkpoint over the six ID mathematical benchmarks and the six OOD benchmarks.
In (c), ``Attn.'' denotes attention projection matrices; ``Gate'', ``Up'', and ``Down'' denote the corresponding MLP matrices; ``MLP'' includes all MLP projection matrices; ``Head+Emb.'' denotes the output head and embedding matrices; and ``All 2D'' denotes all two-dimensional weight matrices.
In (d), ``R'' and ``L'' denote the right and left singular vectors, while ``top'' and ``bottom'' denote the leading and trailing singular subspaces with their dimensions determined using the default threshold \(\rho=0.5\).
``$W_0$'' and ``$\Delta W$'' indicate that the subspace is derived from the pre-SFT weights and the SFT update, respectively.
Shading indicates the default \METHODNAME{} configuration in each panel.
}
    \label{fig:ablation}
\end{figure*}

\subsection{Ablation Study}

We conduct an ablation study on Qwen2.5-7B to examine the sensitivity of \METHODNAME{} to its key hyperparameters and design choices.
All training and evaluation settings follow those in Section~\ref{main_experiments_on_mathematical_reasoning_sft}.
We report the average change relative to the SFT checkpoint separately over the six ID mathematical benchmarks and the six OOD benchmarks.

\paragraph{Hyperparameter sensitivity.}
We first study the sensitivity of \METHODNAME{} to the subspace threshold $\rho$ and calibration strength $\alpha$.
For the latter, we generalize ~\eqref{eq:calibrated_weight} as $W_{\mathrm{cal}}=W_0+\Delta W(I-\alpha P_0)$,
where $\alpha\in[0,1]$ controls the fraction of the SFT update reverted within the selected principal subspace.
Thus, $\alpha=1$ corresponds to the default \METHODNAME{}.
Figure~\ref{fig:ablation}(a) shows that $\rho=0.5$ provides a good balance between ID and OOD performance.
Using a smaller subspace weakens the gains, while a larger one further improves OOD performance at the cost of reduced ID improvement.
We therefore use $\rho=0.5$ as the default.
Figure~\ref{fig:ablation}(b) shows that performance generally improves as the calibration strength increases, with $\alpha=1$ performing best.
Together, these results suggest that \METHODNAME{} is best applied as a hard projection, with $\rho$ as the sole trade-off hyperparameter.

\paragraph{Matrix selection.}
We next examine which weight matrices should be calibrated.
As shown in Figure~\ref{fig:ablation}(c), calibrating attention matrices alone has little effect, whereas jointly calibrating the MLP matrices yields substantially larger improvements.
This is consistent with prior work suggesting that MLP modules serve as important sites for factual knowledge storage and recall~\citep{geva2021transformer,dai2022knowledge,meng2022locating}.
Among individual MLP projections, the Up projection provides the largest gains, while the Down projection alone is ineffective.
Interestingly, attention contributes little on its own but provides additional gains when combined with MLP calibration, suggesting that MLPs drive the primary benefits while attention plays a complementary role.
Finally, applying \METHODNAME{} to all two-dimensional weight matrices gives the strongest overall results.

\paragraph{Subspace selection.}
Finally, we compare different choices of the calibration subspace.
As shown in Figure~\ref{fig:ablation}(d), the leading right singular subspace of the pre-SFT weights, as used by \METHODNAME{}, provides the best balance between ID and OOD improvements.
In contrast, the bottom right singular subspace strongly favors OOD recovery but provides almost no ID gain, indicating a less balanced trade-off.
Moreover, the weaker performance of subspaces derived from $\Delta W$ supports the central premise of Geometric Calibration: the pre-SFT weights provide a more informative reference for refining a completed SFT update.
Overall, the clear advantage of the right over the left singular subspaces is consistent with the input-side formulation of \METHODNAME{}.

\subsection{Experiments on Different Target Domains}

We next examine whether the benefits of \METHODNAME{} extend beyond mathematical reasoning SFT by training on coding and medical data.

\begin{table*}[t]
\centering
\caption{
Experimental results of training Qwen2.5-7B on coding and medical data. ``(M)'' denotes the medical subset. \textbf{Bold} indicates the best result excluding the Base model.
}
\label{tab:different_domains}

\scriptsize
\setlength{\tabcolsep}{2.0pt}

\begin{tabular}{
l
>{\color{gray}}c
c
>{\columncolor{baselinebg}}c
>{\columncolor{baselinebg}}c
>{\columncolor{oursbg}}c
@{\hspace{6pt}}l
>{\color{gray}}c
c
>{\columncolor{baselinebg}}c
>{\columncolor{baselinebg}}c
>{\columncolor{oursbg}}c
}
\toprule

\multicolumn{6}{c}{\textbf{Coding SFT}}
&
\multicolumn{6}{c}{\textbf{Medical SFT}}
\\
\cmidrule(lr){1-6}
\cmidrule(lr){7-12}

\textbf{Benchmark}
& \textbf{Base}
& \textbf{SFT}
& \textbf{DG-Hard}
& \textbf{MonoSoup}
& \textbf{\METHODNAME{}}
&
\textbf{Benchmark}
& \textbf{Base}
& \textbf{SFT}
& \textbf{DG-Hard}
& \textbf{MonoSoup}
& \textbf{\METHODNAME{}}
\\

\midrule

\multicolumn{6}{l}{\textcolor{sectioncolor}{\textit{In-domain (ID), Coding}}}
&
\multicolumn{6}{@{}l}{\textcolor{sectioncolor}{\textit{In-domain (ID), Medical}}}
\\

HumanEval
& 77.44 & \textbf{78.66} & 75.61 & 76.83 & \textbf{78.66}
&
MedMCQA
& 53.74 & 54.79 & 54.22 & 54.12 & \textbf{55.25}
\\

HumanEval+
& 71.34 & 71.95 & 70.73 & 71.34 & \textbf{73.78}
&
MedQA
& 58.68 & \textbf{65.12} & 60.80 & 63.39 & 64.65
\\

MBPP
& 75.40 & 78.04 & 77.78 & \textbf{78.31} & \textbf{78.31}
&
PubMedQA
& 71.70 & 70.00 & 70.60 & 70.00 & \textbf{73.20}
\\

MBPP+
& 65.61 & 67.99 & 67.99 & \textbf{68.78} & 67.99
&
MMLU-Pro (M)
& 55.83 & 57.65 & 55.83 & 59.15 & \textbf{59.54}
\\

LCB v2
& 28.96 & 23.09 & 22.50 & \textbf{24.53} & 22.77
&
GPQA (M)
& 44.87 & 36.92 & 37.95 & 42.82 & \textbf{47.69}
\\

\textbf{Avg. ($\Delta$ vs. SFT)}
& - & -
& \textcolor{negcolor}{-1.02}
& \textcolor{poscolor}{+0.01}
& \textbf{\textcolor{poscolor}{+0.36}}
&
\textbf{Avg. ($\Delta$ vs. SFT)}
& - & -
& \textcolor{negcolor}{-1.02}
& \textcolor{poscolor}{+1.00}
& \textbf{\textcolor{poscolor}{+3.17}}
\\

\midrule

\multicolumn{6}{l}{\textcolor{sectioncolor}{\textit{Out-of-domain (OOD)}}}
&
\multicolumn{6}{@{}l}{\textcolor{sectioncolor}{\textit{Out-of-domain (OOD)}}}
\\

MMLU-Pro
& 40.33 & 38.35 & 39.30 & 40.39 & \textbf{42.32}
&
MMLU-Pro
& 40.33 & \textbf{53.90} & 53.41 & 53.68 & 52.47
\\

GPQA-D
& 26.77 & 25.35 & 22.16 & 26.04 & \textbf{27.27}
&
GPQA-D
& 26.77 & 24.91 & \textbf{26.04} & 25.22 & 25.88
\\

&  &  & 
&  &  & 
LCB v2
& 28.96 & 24.07 & 25.64 & \textbf{27.14} & 26.48
\\

IFEval
& 42.57 & 54.08 & 53.00 & \textbf{54.20} & 53.60
&
IFEval
& 42.57 & 35.01 & 34.41 & 35.97 & \textbf{36.09}
\\

AlpacaEval
& -1.13 & -0.43 & \textbf{-0.23} & -0.25 & -0.39
&
AlpacaEval
& -1.13 & -2.30 & -2.31 & -1.99 & \textbf{-1.73}
\\

HaluEval
& 46.86 & 49.32 & 49.52 & 49.41 & \textbf{50.31}
&
HaluEval
& 46.86 & 63.31 & 64.05 & \textbf{65.39} & 65.10
\\

\textbf{Avg. ($\Delta$ vs. SFT)}
& - & -
& \textcolor{negcolor}{-0.58}
& \textcolor{poscolor}{+0.62}
& \textbf{\textcolor{poscolor}{+1.29}}
&
\textbf{Avg. ($\Delta$ vs. SFT)}
& - & -
& \textcolor{poscolor}{+0.39}
& \textbf{\textcolor{poscolor}{+1.09}}
& \textcolor{poscolor}{+0.90}
\\

\bottomrule
\end{tabular}
\end{table*}

\paragraph{Setup.}
For medical SFT, we fine-tune Qwen2.5-7B on 23,493 high-quality medical reasoning examples from m23k \citep{huang2026m}, as used in prior work~\citep{li2025beyond}. For coding SFT, following prior work~\citep{zhu2026anchored}, we use 100K examples sampled from Magicoder-Evol-Instruct-110K~\citep{wei2024magicoder} as training data, and evaluate ID performance using the EvalPlus library~\citep{liu2023your} on HumanEval, HumanEval+ \citep{chen2021evaluating}, MBPP, and MBPP+ \citep{austin2021program}. LiveCodeBench v2~\citep{jain2025livecodebench} is also included for ID evaluation. The maximum generation length is set to 4,096 tokens. For medical SFT, we use the evaluation toolkit provided by \citet{li2025beyond} and report results on MedMCQA~\citep{pal2022medmcqa}, MedQA~\citep{jin2021disease}, PubMedQA~\citep{jin2019pubmedqa}, MMLU-Pro (Medical)~\citep{wang2024mmlu}, and GPQA (Medical)~\citep{rein2024gpqa}, with a maximum generation length of 8,192 tokens. Additionally, we include DG-Hard and MonoSoup as two strong baselines for comparison. All other training and evaluation settings follow the same protocol as in Section~\ref{main_experiments_on_mathematical_reasoning_sft}.

\paragraph{Results.}
As shown in Table~\ref{tab:different_domains}, \METHODNAME{} remains effective when the training domain shifts to coding or medicine. For coding SFT, where the uncalibrated SFT checkpoint already performs well on most ID benchmarks, \METHODNAME{} largely preserves the learned adaptation while improving OOD performance. For medical SFT, the ID gains are more pronounced, particularly on benchmarks where SFT causes regression relative to the Base model. This pattern suggests that geometric calibration is particularly beneficial when SFT introduces stronger interference with capabilities already encoded in the pre-SFT model. Overall, \METHODNAME{} improves upon SFT across both domains and outperforms two strong baselines, DG-Hard and MonoSoup, demonstrating its applicability across different forms of task adaptation rather than relying on properties specific to mathematical reasoning.

\begin{table*}[t]
\centering
\caption{Compatibility with different SFT methods. ``Original'' denotes the checkpoint produced by each training method. ``Avg. ($\Delta$ vs. Original)'' denotes the average score change relative to the original checkpoint. \textbf{Bold} indicates the best score within each method.}
\label{tab:compatibility_with_different_sft_methods}

\scriptsize
\setlength{\tabcolsep}{3.2pt}

\begin{tabular}{
l
>{\color{gray}}c
c
>{\columncolor{baselinebg}}c
>{\columncolor{oursbg}}c
c
>{\columncolor{baselinebg}}c
>{\columncolor{oursbg}}c
c
>{\columncolor{baselinebg}}c
>{\columncolor{oursbg}}c
}
\toprule

\multirow{2}{*}{\textbf{Benchmark}}
& \multirow{2}{*}{\textbf{Base}}
& \multicolumn{3}{c}{\textbf{DFT}}
& \multicolumn{3}{c}{\textbf{ASFT}}
& \multicolumn{3}{c}{\textbf{RPSFT}}
\\
\cmidrule(lr){3-5}
\cmidrule(lr){6-8}
\cmidrule(lr){9-11}

&
&
\textbf{Original}
& \textbf{MonoSoup}
& \textbf{\METHODNAME{}}
& \textbf{Original}
& \textbf{MonoSoup}
& \textbf{\METHODNAME{}}
& \textbf{Original}
& \textbf{MonoSoup}
& \textbf{\METHODNAME{}}
\\

\midrule

\multicolumn{8}{l}{\textcolor{sectioncolor}{\textit{In-domain (ID)}}} \\

MATH500
& 28.79
& 54.40 & 57.33 & \textbf{58.64}
& 60.55 & \textbf{61.56} & 61.12
& 45.98 & 46.64 & \textbf{48.69}
\\

AIME25
& 0.42
& \textbf{1.88} & 1.46 & 1.25
& 1.67 & \textbf{2.08} & 1.88
& \textbf{1.04} & 0.63 & 0.83
\\

AIME24
& 1.65
& 2.08 & 3.54 & \textbf{3.75}
& \textbf{5.83} & 5.00 & 5.21
& \textbf{1.46} & 1.25 & 1.25
\\

OlympiadBench
& 7.69
& 23.22 & 26.29 & \textbf{27.18}
& 27.98 & \textbf{28.81} & 28.45
& 14.40 & 15.53 & \textbf{17.03}
\\

Minerva Math
& 9.26
& \textbf{20.57} & 20.22 & 20.24
& \textbf{24.43} & 22.43 & 22.73
& 13.01 & 13.81 & \textbf{15.12}
\\

AMC23
& 15.65
& 27.81 & \textbf{29.06} & 28.91
& 35.31 & 36.72 & \textbf{38.28}
& \textbf{21.72} & 17.97 & 20.16
\\

\textbf{Avg. ($\Delta$ vs. Original)}
& -
& - & \textcolor{poscolor}{+1.32} & \textbf{\textcolor{poscolor}{+1.67}}
& - & \textcolor{poscolor}{+0.14} & \textbf{\textcolor{poscolor}{+0.32}}
& - & \textcolor{negcolor}{-0.30} & \textbf{\textcolor{poscolor}{+0.91}}
\\

\midrule

\multicolumn{8}{l}{\textcolor{sectioncolor}{\textit{Out-of-domain (OOD)}}} \\

MMLU-Pro
& 40.33
& \textbf{33.66} & 30.84 & 33.39
& 33.97 & 33.54 & \textbf{34.45}
& 34.67 & 36.38 & \textbf{41.06}
\\

GPQA-D
& 26.77
& 27.27 & \textbf{27.78} & 26.14
& 26.48 & \textbf{27.02} & 25.88
& 24.15 & 25.69 & \textbf{27.27}
\\

LCB v2
& 28.96
& 30.85 & 32.22 & \textbf{33.27}
& 27.14 & \textbf{29.35} & 28.90
& 29.03 & 29.35 & \textbf{29.75}
\\

IFEval
& 42.57
& 43.17 & 46.88 & \textbf{48.20}
& 45.20 & 47.12 & \textbf{48.44}
& 44.48 & 44.48 & \textbf{45.56}
\\

AlpacaEval
& -1.13
& -0.43 & -0.17 & \textbf{-0.13}
& \textbf{-0.08} & -0.09 & -0.09
& -0.03 & \textbf{-0.01} & -0.07
\\

HaluEval
& 46.86
& \textbf{48.04} & 46.97 & 47.11
& 47.06 & 47.43 & \textbf{48.62}
& 46.86 & 47.48 & \textbf{48.68}
\\

\textbf{Avg. ($\Delta$ vs. Original)}
& -
& - & \textcolor{poscolor}{+0.33} & \textbf{\textcolor{poscolor}{+0.90}}
& - & \textcolor{poscolor}{+0.77} & \textbf{\textcolor{poscolor}{+1.07}}
& - & \textcolor{poscolor}{+0.70} & \textbf{\textcolor{poscolor}{+2.18}}
\\

\bottomrule
\end{tabular}
\end{table*}

\subsection{Compatibility with Different SFT Methods}
\label{subsec:compatibility_with_different_sft_methods}

Recently, several variants of SFT have been proposed to improve performance and generalization. We further examine whether \METHODNAME{} is compatible with these methods.

\paragraph{Setup.}
The comparison is conducted on Qwen2.5-7B trained with data from NuminaMath-CoT. Three SFT variants are considered: DFT~\citep{wu2026generalization}, which reweights the token-level objective; ASFT~\citep{zhu2026anchored}, which augments DFT with distributional anchoring; and RPSFT~\citep{jin2026rotation}, which introduces regularization to constrain weight updates during fine-tuning. For each method, we apply \METHODNAME{} to its completed checkpoint using the same pre-SFT checkpoint as the geometric reference. Other training and evaluation settings remain consistent with Section~\ref{main_experiments_on_mathematical_reasoning_sft}.

\paragraph{Results.}
Table~\ref{tab:compatibility_with_different_sft_methods} shows that \METHODNAME{} improves overall ID and OOD performance across all three SFT methods. The consistent gains across methods with different training objectives suggest that the utility of pre-SFT geometry is not tied to a particular SFT update formation process. Notably, the gains are particularly pronounced on RPSFT, even though it already constrains weight geometry during training, suggesting that training-time regularization and post-hoc geometric calibration address complementary aspects.
Moreover, \METHODNAME{} consistently achieves larger aggregate gains than the strong baseline MonoSoup across all three SFT variants. Appendix~\ref{app:compatibility_with_lora} shows that these benefits also extend to LoRA fine-tuning. Overall, these results show that \METHODNAME{} can effectively complement different training objectives by refining their completed updates post hoc.

\subsection{Generalization Across Model Families and Scales}
\label{subsec:generalization_across_models_families_and_scales}

We further evaluate whether \METHODNAME{} generalizes across different model families and scales, covering Qwen3.5, Qwen3, Olmo-3, and Llama-3.1 models from 2B to 8B parameters.

\paragraph{Setup.}
The experiments use Qwen3.5-2B-Base, Qwen3-8B-Base, Olmo-3-7B-Instruct-SFT~\citep{olmo2025olmo} and Llama-3.1-8B-Instruct~\citep{grattafiori2024llama} as starting checkpoints for mathematical SFT. Math-CoT-20k is employed as the training data, which contains longer reasoning trajectories and enables more substantial performance improvements through SFT. Other training and evaluation settings are kept the same as in Section~\ref{main_experiments_on_mathematical_reasoning_sft}.

\paragraph{Results.}
As shown in Table~\ref{tab:generalization_across_models_families_and_scales}, \METHODNAME{} yields positive average gains in both ID and OOD performance across all evaluated models. Importantly, the OOD improvements are accompanied by further ID gains across different families and scales, suggesting that geometric calibration refines rather than reverses the SFT adaptation. The gains are particularly pronounced on Llama-3.1-8B, where SFT causes substantial OOD capability regression; in this setting, \METHODNAME{} improves every evaluated OOD benchmark while further strengthening ID performance. Even at the smallest 2B scale, \METHODNAME{} still performs well, indicating that its benefits are not limited to larger models. Additional results in Appendix~\ref{app:additional_model_family_scale_results} show consistent gains on Qwen3-4B-Base and larger average improvements than MonoSoup across all evaluated models. Overall, these results demonstrate the broad applicability of \METHODNAME{} across model families and scales.

\begin{table*}[t]
\centering
\caption{Generalization of \METHODNAME{} across different model families and scales.}
\label{tab:generalization_across_models_families_and_scales}

\scriptsize
\setlength{\tabcolsep}{4.5pt}

\begin{tabular}{
l
>{\color{gray}}c
c
>{\columncolor{oursbg}}c
>{\color{gray}}c
c
>{\columncolor{oursbg}}c
>{\color{gray}}c
c
>{\columncolor{oursbg}}c
>{\color{gray}}c
c
>{\columncolor{oursbg}}c
}
\toprule

\multirow{2}{*}{\textbf{Benchmark}}
& \multicolumn{3}{c}{\textbf{Qwen3.5-2B-Base}}
& \multicolumn{3}{c}{\textbf{Qwen3-8B-Base}}
& \multicolumn{3}{c}{\textbf{Olmo-3-7B-Instruct-SFT}}
& \multicolumn{3}{c}{\textbf{Llama-3.1-8B-Instruct}}
\\
\cmidrule(lr){2-4}
\cmidrule(lr){5-7}
\cmidrule(lr){8-10}
\cmidrule(lr){11-13}

&
\textbf{Base}
& \textbf{SFT}
& \textbf{\METHODNAME{}}
&
\textbf{Base}
& \textbf{SFT}
& \textbf{\METHODNAME{}}
&
\textbf{Base}
& \textbf{SFT}
& \textbf{\METHODNAME{}}
&
\textbf{Base}
& \textbf{SFT}
& \textbf{\METHODNAME{}}
\\

\midrule

\multicolumn{7}{l}{\textcolor{sectioncolor}{\textit{In-domain (ID)}}} \\

MATH500
& 72.40 & 76.47 & \textbf{78.00}
& 55.93 & 83.73 & \textbf{84.87}
& 60.80 & 87.93 & \textbf{89.33}
& 39.07 & 49.20 & \textbf{53.33}
\\

AIME25
& 15.63 & 17.29 & \textbf{17.50}
& 4.58 & 23.96 & \textbf{24.38}
& 3.33 & \textbf{35.00} & 34.38
& 0.21 & \textbf{1.46} & 1.25
\\

AIME24
& 12.92 & \textbf{16.67} & 15.21
& 7.29 & \textbf{22.71} & \textbf{22.71}
& 6.67 & 41.25 & \textbf{41.46}
& 1.88 & 1.67 & \textbf{2.29}
\\

OlympiadBench
& 37.09 & 40.15 & \textbf{41.38}
& 24.05 & 46.86 & \textbf{48.79}
& 25.88 & 55.26 & \textbf{56.79}
& 10.47 & 18.96 & \textbf{21.04}
\\

Minerva Math
& 27.08 & 39.95 & \textbf{40.32}
& 19.49 & \textbf{41.79} & 40.44
& 25.74 & 33.95 & \textbf{34.56}
& 15.56 & 18.01 & \textbf{21.57}
\\

AMC23
& 49.22 & 53.59 & \textbf{55.16}
& 32.50 & 64.22 & \textbf{67.34}
& 37.03 & 79.53 & \textbf{82.19}
& 14.69 & 26.09 & \textbf{30.16}
\\

\textbf{Avg. (\(\Delta\) vs. SFT)}
& - & - & \textcolor{poscolor}{\textbf{+0.58}}
& - & - & \textcolor{poscolor}{\textbf{+0.88}}
& - & - & \textcolor{poscolor}{\textbf{+0.97}}
& - & - & \textcolor{poscolor}{\textbf{+2.38}}
\\

\midrule

\multicolumn{7}{l}{\textcolor{sectioncolor}{\textit{Out-of-domain (OOD)}}} \\

MMLU-Pro
& 46.40 & 52.60 & \textbf{54.00}
& 56.70 & \textbf{56.40} & 56.20
& 42.70 & 44.40 & \textbf{44.90}
& 43.50 & 40.80 & \textbf{44.80}
\\

GPQA-D
& 30.64 & 29.97 & \textbf{32.66}
& 36.36 & \textbf{29.46} & \textbf{29.46}
& 28.79 & 22.22 & \textbf{25.93}
& 23.74 & 15.49 & \textbf{19.87}
\\

LCB v2
& 14.94 & 11.55 & \textbf{13.31}
& 33.66 & 24.92 & \textbf{29.29}
& 17.09 & \textbf{22.44} & 21.14
& 18.92 & 8.41 & \textbf{12.20}
\\

IFEval
& 57.67 & 45.32 & \textbf{53.84}
& 54.08 & 35.85 & \textbf{40.89}
& 83.57 & 79.26 & \textbf{82.49}
& 81.89 & 29.14 & \textbf{36.81}
\\

AlpacaEval
& 0.21 & -0.84 & \textbf{-0.23}
& -0.39 & -2.32 & \textbf{-1.10}
& 0.28 & 0.34 & \textbf{0.38}
& 1.17 & -1.87 & \textbf{-0.38}
\\

HaluEval
& 50.00 & \textbf{59.97} & 59.20
& 49.33 & 72.13 & \textbf{72.30}
& 54.23 & \textbf{55.47} & 55.20
& 62.73 & 63.37 & \textbf{63.97}
\\

\textbf{Avg. (\(\Delta\) vs. SFT)}
& - & - & \textcolor{poscolor}{\textbf{+2.37}}
& - & - & \textcolor{poscolor}{\textbf{+1.77}}
& - & - & \textcolor{poscolor}{\textbf{+0.99}}
& - & - & \textcolor{poscolor}{\textbf{+3.66}}
\\

\bottomrule
\end{tabular}
\end{table*}

\subsection{Additional Analyses}
Further analyses show that \METHODNAME{} is effective at intermediate SFT
checkpoints (Appendix~\ref{app:intermediate_sft_checkpoints}), robust to SFT
training hyperparameters (Appendix~\ref{app:robustness_sft_training_hyperparameters}),
and more effective as a post-hoc calibration than as a training-time projection
(Appendix~\ref{app:training_time_vs_post_hoc_projection}).

%% file: 6-conclusion.tex
\section{Conclusion}

We study \mbox{\emph{Geometric Calibration}}, a perspective for refining SFT updates using the geometry of the pre-SFT weights as a reference. We develop \METHODNAME{} as a lightweight post-hoc implementation of this perspective, which removes SFT update components within the principal input subspace of the pre-SFT weights while preserving the remaining adaptation. \METHODNAME{} operates directly on the pre- and post-SFT checkpoints, with no data or additional training required. Overall, across multiple model families, scales, training domains, and SFT variants, \METHODNAME{} improves both ID and OOD performance over the SFT checkpoint and outperforms representative post-hoc baselines. Our ablations further support the principal right singular subspace of the pre-SFT weights as an effective calibration reference. Taken together, these results highlight the value of pre-SFT weight geometry for refining completed adaptations and, more broadly, motivate \mbox{\emph{Geometric Calibration}} as a general principle for continued model improvement beyond SFT.

%% file: 7-statements.tex
\subsection*{AI use statement}

In this work, we used generative AI tools to assist with language editing and polishing, literature search and discovery, and software code assistance. All identified literature was independently reviewed and verified by the authors. Regarding code assistance, generative AI tools were used only to assist with code editing, completion, and debugging. All AI-assisted code was manually reviewed, tested, and validated. The research ideas, methodology, experimental design, code structure and implementation logic, and interpretation of the results were developed by the authors. All AI-assisted content was validated by the authors. We take responsibility for the final content of this work, including text, claims or artifacts produced with the aid of generative AI.

\subsection*{Ethics statement}

Our study does not involve human-subject experiments or the collection of personal or sensitive data. The experiments are conducted using publicly available models, datasets, and standard evaluation benchmarks. Since \METHODNAME{} operates on existing language-model checkpoints, the resulting models may retain safety risks associated with the underlying LLMs, including biased, inaccurate, or potentially harmful outputs. Appropriate validation, monitoring, and task-specific safeguards should therefore be maintained in downstream applications.

\subsection*{Reproducibility statement}

Detailed training settings, evaluation protocols, and baseline details are provided in Appendices~\ref{app:detailed_training_settings_of_the_main_experiments}, \ref{app:detailed_evaluation_settings_of_the_main_experiments}, and \ref{app:baseline_details}. Additional experiments examine robustness across models, SFT settings, and training configurations. Our source code is available at: \url{https://github.com/jianzhubao/SPACE}.

\subsubsection*{Acknowledgments}

This research was conducted at the AUMOVIO-NTU Corporate Lab.
This research/project is supported by A*STAR under the RIE2025 Industry Alignment Fund – Industry Collaboration Projects (IAF-ICP) Funding Initiative (Award: I2501E0045), as well as cash and in-kind contribution from the industry partner(s).

%% file: Appendix.tex
\etocsetlocaltop.toc{part}

\etocsetnexttocdepth{subsection}

\etocsettocstyle{
    \vspace{0.5em}
    \noindent\textbf{\Large Appendix Contents}\par
    \vspace{0.8em}
}{
    \vspace{0.8em}
}

\localtableofcontents

\section{Derivation of the Principal Input Projector}
\label{app:principal_input_projector}

We derive the solution to the projection problem in
\eqref{eq:principal_reference}. Specifically, we solve the following optimization problem:
\begin{align*}
P_0
\in
\underset{\substack{
P=P^\top=P^2\\
\operatorname{rank}(P)=k
}}{\arg\min}
\|W_0-W_0P\|_F^2.
\end{align*}

Since \(P=P^\top=P^2\),
\begin{align*}
\|W_0-W_0P\|_F^2
&=
\|W_0(I-P)\|_F^2 \\
&=
\operatorname{tr}\!\left(
(I-P)W_0^\top W_0(I-P)
\right) \\
&=
\operatorname{tr}(W_0^\top W_0)
-
\operatorname{tr}(PW_0^\top W_0).
\end{align*}

Therefore, minimizing \eqref{eq:principal_reference}
is equivalent to maximizing
\[
\operatorname{tr}(PW_0^\top W_0)
\]
over rank-\(k\) orthogonal projectors.

Writing
\[
W_0=U_0\Sigma_0V_0^\top,
\]
we have
\[
W_0^\top W_0
=
V_0\Sigma_0^2V_0^\top.
\]

Any rank-\(k\) orthogonal projector can be written as
\[
P=QQ^\top,
\qquad
Q^\top Q=I_k.
\]
Therefore,
\[
\operatorname{tr}(PW_0^\top W_0)
=
\operatorname{tr}(Q^\top W_0^\top W_0 Q).
\]

By the Ky Fan maximum principle~\citep{fan1949theorem},
\[
\max_{Q^\top Q=I_k}
\operatorname{tr}(Q^\top W_0^\top W_0 Q)
=
\sum_{i=1}^k \sigma_{0,i}^2,
\]
and the maximum is attained when the columns of \(Q\) span the
eigenspace associated with the \(k\) largest eigenvalues of
\(W_0^\top W_0\). Since these eigenvectors are precisely the leading
\(k\) right singular vectors of \(W_0\), we may choose
\[
Q=V_{0,k}.
\]
Hence, an optimal projector is
\[
P_0=V_{0,k}V_{0,k}^\top.
\]

\section{Detailed Experimental Settings}

\subsection{Detailed Training Settings of the Main Experiments}
\label{app:detailed_training_settings_of_the_main_experiments}

Unless otherwise specified, we use AdamW with a global batch size of 256 and a learning rate of \(5\times10^{-5}\). The learning rate follows a cosine decay schedule with a warm-up ratio of \(0.1\), and all models are trained for one epoch. We use standard next-token prediction over the response tokens as the SFT objective. All experiments are conducted using four NVIDIA H200 GPUs.

\subsection{Detailed Evaluation Settings of the Main Experiments}
\label{app:detailed_evaluation_settings_of_the_main_experiments}

Following prior studies \citep{wu2026generalization,zhu2026anchored}, for the mathematical benchmarks (MATH-500, Minerva Math, OlympiadBench, AIME 2024, AIME 2025, and AMC 2023), we adopt the Qwen2.5-Math \citep{yang2024qwen2} evaluation pipeline and use the default CoT prompt template with a temperature of \(1.0\) and top-\(p\) of \(1.0\). Following the benchmark suite used by \citet{ren2026rethinking}, we further consider OOD benchmarks. In particular, we use MMLU-Pro \citep{wang2024mmlu} for broad knowledge-intensive reasoning, GPQA-Diamond (GPQA-D) \citep{rein2024gpqa} for graduate-level scientific reasoning, LiveCodeBench v2 (LCB v2) \citep{jain2025livecodebench} for code reasoning, IFEval \citep{zhou2023instruction} for instruction following, AlpacaEval (RM) \citep{dubois2024length} for open-ended response quality, and HaluEval \citep{li2023halueval} for hallucination detection. For these evaluations, we follow the prompt templates and evaluation settings of \citet{ren2026rethinking}, using a temperature of \(0.6\) and top-\(p\) of \(0.95\).

The remaining evaluation settings differ slightly between Qwen2.5-7B and Qwen3-14B-Base, as their training data have substantially different maximum CoT lengths (approximately 3K vs. 16K tokens), leading to different inference costs. For Qwen2.5-7B, we set the maximum generation length to 4,096 tokens. For Qwen3-14B-Base, we set the maximum generation length to 16,384 tokens. Moreover, for Qwen2.5-7B, we report avg@16 on the mathematical benchmarks and GPQA-Diamond, and evaluate MMLU-Pro and HaluEval on their full test sets. For Qwen3-14B-Base, following \citet{ren2026rethinking}, we report avg@3 on MATH-500, Minerva Math, OlympiadBench, and GPQA-Diamond, and avg@16 on the remaining mathematical benchmarks. Also, MMLU-Pro and HaluEval are evaluated on subsets of 1K and 3K examples. For both models, we report pass@1 for MMLU-Pro and HaluEval, strict instruction-level accuracy for IFEval, and avg@3 for LiveCodeBench v2. For AlpacaEval (RM), we generate one response for each instruction and report the average reward score from the \texttt{allenai/Llama-3.1-8B-Instruct-RM-RB2} reward model.\footnote{\url{https://huggingface.co/allenai/Llama-3.1-8B-Instruct-RM-RB2}}

\subsection{Baseline Details}
\label{app:baseline_details}

We compare \METHODNAME{} with the following representative post-hoc model editing/merging baselines. 

\begin{itemize}
    \item WiSE-FT~\citep{wortsman2022robust} linearly interpolates the pre-SFT and SFT checkpoints.
    \item TIES~\citep{yadav2023ties} applies task-vector trimming and sign-aware merging.
    \item LiNeS~\citep{wang2025lines} applies depth-dependent layer-wise scaling to the fine-tuning update.
    \item DARE~\citep{yu2024language} sparsifies the fine-tuning update through random dropping and rescaling.
    \item POME~\citep{liu2025pome} refines the fine-tuning update through Muon-style projection.
    \item FAPM~\citep{huang2025mitigating} performs forgetting-aware pruning of the fine-tuning update.
    \item DG-Hard~\citep{abro2026spectral} performs spectral filtering of the fine-tuning update.
    \item MonoSoup~\citep{abdollahpoorrostam2026model} decomposes and reweights spectral components of the fine-tuning update.
\end{itemize}

\paragraph{Implementation details.}
We follow the recommended settings in the original papers or their released implementations, with the following adaptations.
For TIES, we use a single SFT update, so the method reduces to global magnitude trimming.
For FAPM, the default setting performs poorly, likely due to excessive update pruning in our setting. With the scoring rule unchanged, we evaluate several retention ratios and choose to retain $90\%$ of entries per tensor, which provides good overall performance.

\paragraph{Comparison with \METHODNAME{}.}
WiSE-FT and LiNeS scale updates globally or by layer depth.
TIES, DARE, and FAPM prune individual update entries using magnitude-based scores or random masking.
Although some of these methods incorporate pre-SFT weight information, they do not explicitly account for the geometry of the pre-SFT weights.
POME, DG-Hard, and MonoSoup exploit spectral structure, but define their transformations using the learned update's own singular components.
\METHODNAME{} instead calibrates post-SFT checkpoints against the geometry of the pre-SFT weights, removing update components within their principal input subspaces while preserving the update on the orthogonal complements.

\section{Additional Experiments}
\label{app:additional_experiments}

\subsection{Compatibility with LoRA Fine-Tuning}
\label{app:compatibility_with_lora}

We further examine whether \METHODNAME{} is compatible with parameter-efficient fine-tuning. Specifically, we fine-tune Qwen2.5-7B with LoRA on NuminaMath-CoT using rank $r=64$ and scaling factor $\alpha=128$, while keeping the remaining training and evaluation settings consistent with Section~\ref{main_experiments_on_mathematical_reasoning_sft}. We then apply \METHODNAME{} to the resulting LoRA checkpoint using the same pre-SFT checkpoint as the geometric reference. As shown in Table~\ref{tab:compatibility_with_lora}, \METHODNAME{} improves average ID and OOD performance over the original LoRA checkpoint, with particularly pronounced OOD gains. It also achieves larger aggregate gains than MonoSoup. These results suggest that the applicability of \METHODNAME{} extends beyond full-parameter SFT to parameter-efficient adaptation.

\begin{table*}[t]
\centering
\caption{Compatibility of \METHODNAME{} with LoRA fine-tuning on Qwen2.5-7B ($r=64$, $\alpha=128$). ``Original'' denotes the LoRA fine-tuned checkpoint. ``Avg. ($\Delta$ vs. Original)'' denotes the average score change after applying MonoSoup or \METHODNAME{} relative to Original within each benchmark group. \textbf{Bold} indicates the best score.}
\label{tab:compatibility_with_lora}

\scriptsize

\begin{tabular}{
l
c
>{\columncolor{baselinebg}}c
>{\columncolor{oursbg}}c
}
\toprule

\multirow{2}{*}{\textbf{Benchmark}}
& \multicolumn{3}{c}{\textbf{LoRA}}
\\
\cmidrule(lr){2-4}

&
\textbf{Original}
& \textbf{MonoSoup}
& \textbf{\METHODNAME{}}
\\

\midrule

\multicolumn{4}{l}{\textcolor{sectioncolor}{\textit{In-domain (ID)}}} \\

MATH500
& 45.30 & 46.06 & \textbf{46.34}
\\

AIME25
& 0.21 & \textbf{1.25} & 0.63
\\

AIME24
& 0.83 & 0.83 & \textbf{2.50}
\\

OlympiadBench
& 14.82 & 14.95 & \textbf{15.40}
\\

Minerva Math
& 12.94 & 13.03 & \textbf{13.63}
\\

AMC23
& \textbf{20.16} & 17.97 & 18.91
\\

\textbf{Avg. ($\Delta$ vs. Original)}
& - & \textcolor{negcolor}{-0.03}
& \textbf{\textcolor{poscolor}{+0.53}}
\\

\midrule

\multicolumn{4}{l}{\textcolor{sectioncolor}{\textit{Out-of-domain (OOD)}}} \\

MMLU-Pro
& 38.20 & 37.28 & \textbf{43.28}
\\

GPQA-D
& 22.35 & 22.85 & \textbf{27.84}
\\

LCB v2
& 27.98 & \textbf{28.25} & 27.98
\\

IFEval
& 42.81 & 44.00 & \textbf{48.44}
\\

AlpacaEval
& -0.08 & \textbf{-0.06} & -0.12
\\

HaluEval
& 46.75 & 46.87 & \textbf{51.22}
\\

\textbf{Avg. ($\Delta$ vs. Original)}
& - & \textcolor{poscolor}{+0.20}
& \textbf{\textcolor{poscolor}{+3.44}}
\\

\bottomrule
\end{tabular}
\end{table*}

\subsection{Additional Results on Model Families and Scales}
\label{app:additional_model_family_scale_results}

We provide additional results for the model generalization experiments in Section~\ref{subsec:generalization_across_models_families_and_scales}. In addition to the models reported in the main text, we include Qwen3-4B-Base and compare \METHODNAME{} with MonoSoup across all evaluated models. Tables~\ref{tab:generalization_qwen_scales} and~\ref{tab:generalization_olmo_llama} show that \METHODNAME{} consistently improves the average ID and OOD performance of the corresponding SFT checkpoints across all evaluated models. The gains remain consistent across the evaluated Qwen scales and extend to the Olmo and Llama model families. Moreover, \METHODNAME{} achieves larger average improvements than MonoSoup on both ID and OOD benchmarks for every evaluated model, further demonstrating its applicability.

\begin{table*}[t]
\centering
\caption{
Generalization of \METHODNAME{} across Qwen models at different scales (2B, 4B, and 8B).
}
\label{tab:generalization_qwen_scales}

\scriptsize
\setlength{\tabcolsep}{2.0pt}

\begin{tabular}{
l
>{\color{gray}}c
c
>{\columncolor{baselinebg}}c
>{\columncolor{oursbg}}c
>{\color{gray}}c
c
>{\columncolor{baselinebg}}c
>{\columncolor{oursbg}}c
>{\color{gray}}c
c
>{\columncolor{baselinebg}}c
>{\columncolor{oursbg}}c
}
\toprule

\multirow{2}{*}{\textbf{Benchmark}}
& \multicolumn{4}{c}{\textbf{Qwen3.5-2B-Base}}
& \multicolumn{4}{c}{\textbf{Qwen3-4B-Base}}
& \multicolumn{4}{c}{\textbf{Qwen3-8B-Base}}
\\
\cmidrule(lr){2-5}
\cmidrule(lr){6-9}
\cmidrule(lr){10-13}

&
\textbf{Base}
& \textbf{SFT}
& \textbf{MonoSoup}
& \textbf{\METHODNAME{}}
&
\textbf{Base}
& \textbf{SFT}
& \textbf{MonoSoup}
& \textbf{\METHODNAME{}}
&
\textbf{Base}
& \textbf{SFT}
& \textbf{MonoSoup}
& \textbf{\METHODNAME{}}
\\

\midrule

\multicolumn{13}{l}{\textcolor{sectioncolor}{\textit{In-domain (ID)}}} \\

MATH500
& 72.40 & 76.47 & \textbf{78.07} & 78.00
& 40.13 & 78.73 & 80.53 & \textbf{81.33}
& 55.93 & 83.73 & 84.67 & \textbf{84.87}
\\

AIME25
& 15.63 & 17.29 & 17.29 & \textbf{17.50}
& 2.29 & 17.50 & \textbf{20.00} & 18.13
& 4.58 & 23.96 & 23.96 & \textbf{24.38}
\\

AIME24
& 12.92 & \textbf{16.67} & 14.79 & 15.21
& 5.63 & 17.71 & 16.25 & \textbf{18.54}
& 7.29 & 22.71 & \textbf{23.33} & 22.71
\\

OlympiadBench
& 37.09 & 40.15 & \textbf{41.93} & 41.38
& 19.56 & 40.89 & \textbf{43.26} & 42.72
& 24.05 & 46.86 & 48.49 & \textbf{48.79}
\\

Minerva Math
& 27.08 & 39.95 & \textbf{40.81} & 40.32
& 16.79 & 37.99 & 36.76 & \textbf{39.22}
& 19.49 & 41.79 & \textbf{42.28} & 40.44
\\

AMC23
& 49.22 & 53.59 & 54.53 & \textbf{55.16}
& 22.03 & 56.41 & 57.19 & \textbf{59.69}
& 32.50 & 64.22 & 65.47 & \textbf{67.34}
\\

\textbf{Avg. ($\Delta$ vs. SFT)}
& - & - & \textcolor{poscolor}{+0.55} & \textcolor{poscolor}{\textbf{+0.58}}
& - & - & \textcolor{poscolor}{+0.79} & \textcolor{poscolor}{\textbf{+1.73}}
& - & - & \textcolor{poscolor}{+0.82} & \textcolor{poscolor}{\textbf{+0.88}}
\\

\midrule

\multicolumn{13}{l}{\textcolor{sectioncolor}{\textit{Out-of-domain (OOD)}}} \\

MMLU-Pro
& 46.40 & 52.60 & 51.80 & \textbf{54.00}
& 37.10 & 47.30 & 48.10 & \textbf{49.20}
& 56.70 & 56.40 & \textbf{56.50} & 56.20
\\

GPQA-D
& 30.64 & 29.97 & \textbf{32.83} & 32.66
& 18.86 & 21.04 & 20.54 & \textbf{21.72}
& 36.36 & \textbf{29.46} & 28.96 & \textbf{29.46}
\\

LCB v2
& 14.94 & 11.55 & 11.35 & \textbf{13.31}
& 31.83 & 15.33 & 17.87 & \textbf{18.33}
& 33.66 & 24.92 & 27.66 & \textbf{29.29}
\\

IFEval
& 57.67 & 45.32 & 51.80 & \textbf{53.84}
& 46.76 & 30.46 & 31.65 & \textbf{32.61}
& 54.08 & 35.85 & 37.17 & \textbf{40.89}
\\

AlpacaEval
& 0.21 & -0.84 & -0.52 & \textbf{-0.23}
& -2.33 & -3.08 & -2.83 & \textbf{-2.71}
& -0.39 & -2.32 & -1.75 & \textbf{-1.10}
\\

HaluEval
& 50.00 & \textbf{59.97} & 59.87 & 59.20
& 48.50 & 55.13 & 55.70 & \textbf{57.17}
& 49.33 & 72.13 & \textbf{74.00} & 72.30
\\

\textbf{Avg. ($\Delta$ vs. SFT)}
& - & - & \textcolor{poscolor}{+1.43} & \textcolor{poscolor}{\textbf{+2.37}}
& - & - & \textcolor{poscolor}{+0.81} & \textcolor{poscolor}{\textbf{+1.69}}
& - & - & \textcolor{poscolor}{+1.02} & \textcolor{poscolor}{\textbf{+1.77}}
\\

\bottomrule
\end{tabular}
\end{table*}

\begin{table*}[t]
\centering
\caption{Generalization of \METHODNAME{} across Olmo and Llama model families.}
\label{tab:generalization_olmo_llama}

\scriptsize

\begin{tabular}{
l
>{\color{gray}}c
c
>{\columncolor{baselinebg}}c
>{\columncolor{oursbg}}c
>{\color{gray}}c
c
>{\columncolor{baselinebg}}c
>{\columncolor{oursbg}}c
}
\toprule

\multirow{2}{*}{\textbf{Benchmark}}
& \multicolumn{4}{c}{\textbf{Olmo-3-7B-Instruct-SFT}}
& \multicolumn{4}{c}{\textbf{Llama-3.1-8B-Instruct}}
\\
\cmidrule(lr){2-5}
\cmidrule(lr){6-9}

&
\textbf{Base}
& \textbf{SFT}
& \textbf{MonoSoup}
& \textbf{\METHODNAME{}}
&
\textbf{Base}
& \textbf{SFT}
& \textbf{MonoSoup}
& \textbf{\METHODNAME{}}
\\

\midrule

\multicolumn{9}{l}{\textcolor{sectioncolor}{\textit{In-domain (ID)}}} \\

MATH500
& 60.80 & 87.93 & 88.67 & \textbf{89.33}
& 39.07 & 49.20 & 52.80 & \textbf{53.33}
\\

AIME25
& 3.33 & \textbf{35.00} & 30.83 & 34.38
& 0.21 & \textbf{1.46} & 1.04 & 1.25
\\

AIME24
& 6.67 & 41.25 & 38.33 & \textbf{41.46}
& 1.88 & 1.67 & \textbf{2.50} & 2.29
\\

OlympiadBench
& 25.88 & 55.26 & 55.75 & \textbf{56.79}
& 10.47 & 18.96 & 19.46 & \textbf{21.04}
\\

Minerva Math
& 25.74 & 33.95 & 34.19 & \textbf{34.56}
& 15.56 & 18.01 & 20.47 & \textbf{21.57}
\\

AMC23
& 37.03 & 79.53 & 80.94 & \textbf{82.19}
& 14.69 & 26.09 & 27.50 & \textbf{30.16}
\\

\textbf{Avg. ($\Delta$ vs. SFT)}
& - & - & \textcolor{negcolor}{-0.70} & \textcolor{poscolor}{\textbf{+0.97}}
& - & - & \textcolor{poscolor}{+1.40} & \textcolor{poscolor}{\textbf{+2.38}}
\\

\midrule

\multicolumn{9}{l}{\textcolor{sectioncolor}{\textit{Out-of-domain (OOD)}}} \\

MMLU-Pro
& 42.70 & 44.40 & 43.20 & \textbf{44.90}
& 43.50 & 40.80 & 44.40 & \textbf{44.80}
\\

GPQA-D
& 28.79 & 22.22 & 25.59 & \textbf{25.93}
& 23.74 & 15.49 & 18.86 & \textbf{19.87}
\\

LCB v2
& 17.09 & \textbf{22.44} & 21.98 & 21.14
& 18.92 & 8.41 & 11.61 & \textbf{12.20}
\\

IFEval
& 83.57 & 79.26 & 78.90 & \textbf{82.49}
& 81.89 & 29.14 & 34.65 & \textbf{36.81}
\\

AlpacaEval
& 0.28 & 0.34 & 0.34 & \textbf{0.38}
& 1.17 & -1.87 & -0.83 & \textbf{-0.38}
\\

HaluEval
& 54.23 & 55.47 & \textbf{55.83} & 55.20
& 62.73 & 63.37 & 63.77 & \textbf{63.97}
\\

\textbf{Avg. ($\Delta$ vs. SFT)}
& - & - & \textcolor{poscolor}{+0.29} & \textcolor{poscolor}{\textbf{+0.99}}
& - & - & \textcolor{poscolor}{+2.85} & \textcolor{poscolor}{\textbf{+3.66}}
\\

\bottomrule
\end{tabular}
\end{table*}

\subsection{Performance at Intermediate SFT Checkpoints}
\label{app:intermediate_sft_checkpoints}

\paragraph{Setup.}
We further examine whether \METHODNAME{} remains effective when applied before the end of SFT. For the Qwen2.5-7B mathematical SFT experiment, one epoch corresponds to approximately 390 optimization steps. We save intermediate checkpoints at 100, 200, and 300 steps along the same training trajectory, in addition to the final 390-step checkpoint. \METHODNAME{} is applied independently to each checkpoint using the same default configuration as in the main experiments. We evaluate all checkpoints on the same six ID and six OOD benchmarks described in Section~\ref{main_experiments_on_mathematical_reasoning_sft}. For clarity, we report the average performance change relative to the final 390-step SFT checkpoint.

\begin{figure}[t]
    \centering
    \includegraphics[width=0.65\textwidth]{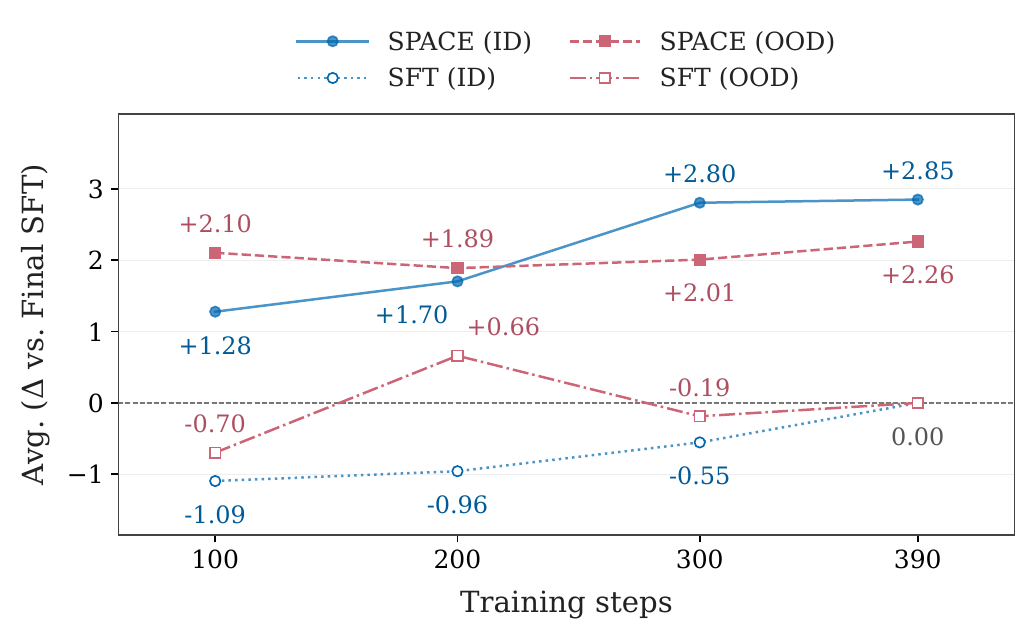}
    \caption{
    Performance of SFT and \METHODNAME{} at different stages of the Qwen2.5-7B training
    trajectory. We report the average change over the six ID and six OOD benchmarks
    relative to the final 390-step SFT checkpoint. The 390-step checkpoint corresponds
    to one epoch of SFT.
    }
    \label{fig:intermediate_sft_checkpoints}
\end{figure}

\paragraph{Results.}
As shown in Figure~\ref{fig:intermediate_sft_checkpoints}, the benefits of \METHODNAME{} emerge early during SFT. After only 100 training steps, \METHODNAME{} already outperforms the final 390-step SFT checkpoint by 1.28 points on ID benchmarks and 2.10 points on OOD benchmarks, whereas the corresponding 100-step SFT checkpoint remains 1.09 and 0.70 points below the final SFT checkpoint, respectively. As training proceeds, the ID improvement from \METHODNAME{} further increases, reaching 2.85 points at 390 steps, while the OOD improvement remains relatively stable throughout training. These results show that \METHODNAME{} does not rely on a fully trained SFT checkpoint and can provide substantial gains even at an early stage of adaptation.

\subsection{Robustness to SFT Training Hyperparameters}
\label{app:robustness_sft_training_hyperparameters}

\paragraph{Setup.}
We further examine the robustness of \METHODNAME{} to different SFT training hyperparameters. Using Qwen2.5-7B fine-tuned on NuminaMath-CoT, we vary the batch size, learning rate, and number of training epochs while keeping the remaining settings fixed. For all experiments, \METHODNAME{} uses the same default configuration as in the main experiments.

\paragraph{Results.}
Tables~\ref{tab:qwen25_numina_batch_size}, \ref{tab:qwen25_numina_learning_rate}, and~\ref{tab:qwen25_numina_epochs} show that \METHODNAME{} yields positive average gains over the corresponding SFT checkpoints across all evaluated training hyperparameters. Although the magnitude of the gains varies across settings, the improvements consistently extend to both ID and OOD performance. These results suggest that \METHODNAME{} is robust to variations in SFT optimization, rather than being specific to a particular choice of training hyperparameters.

\begin{table*}[t]
\centering
\caption{Performance of \METHODNAME{} across SFT batch sizes for Qwen2.5-7B fine-tuned on NuminaMath-CoT. The learning rate is fixed at $5\times10^{-5}$, and training lasts one epoch. Projection uses $\rho=0.5$, $\alpha=1$, and all weight matrices. Bold marks the best score in each pair. $^{\dagger}$ denotes the hyperparameters we used in the main experiment.}
\label{tab:qwen25_numina_batch_size}
\scriptsize

\begin{tabular}{l*{3}{c>{\columncolor{oursbg}}c}}
\toprule
\multirow{2}{*}{\textbf{Benchmark}} & \multicolumn{2}{c}{\textbf{Batch size 128}} & \multicolumn{2}{c}{\textbf{Batch size 256}$^{\dagger}$} & \multicolumn{2}{c}{\textbf{Batch size 512}} \\
\cmidrule(lr){2-3} \cmidrule(lr){4-5} \cmidrule(lr){6-7}
 & \textbf{SFT} & \textbf{\METHODNAME{}} & \textbf{SFT} & \textbf{\METHODNAME{}} & \textbf{SFT} & \textbf{\METHODNAME{}} \\
\midrule
\multicolumn{7}{l}{\textcolor{sectioncolor}{\textit{In-domain (ID)}}} \\
MATH500 & 43.40 & \textbf{48.88} & 44.84 & \textbf{49.36} & 45.31 & \textbf{48.45} \\
AIME25 & 0.83 & \textbf{1.04} & 0.83 & \textbf{1.46} & 0.42 & \textbf{0.83} \\
AIME24 & 0.21 & \textbf{2.08} & 1.04 & \textbf{1.46} & 0.63 & \textbf{1.25} \\
OlympiadBench & 13.06 & \textbf{18.49} & 14.08 & \textbf{17.72} & 14.31 & \textbf{17.70} \\
Minerva Math & 12.39 & \textbf{15.72} & 12.68 & \textbf{15.26} & 12.71 & \textbf{14.36} \\
AMC23 & 17.34 & \textbf{25.31} & 19.53 & \textbf{24.84} & 18.44 & \textbf{22.34} \\

\textbf{Avg. ($\Delta$ vs. SFT)} & - & \textcolor{poscolor}{\textbf{+4.05}} & - & \textcolor{poscolor}{\textbf{+2.85}} & - & \textcolor{poscolor}{\textbf{+2.19}} \\
\midrule
\multicolumn{7}{l}{\textcolor{sectioncolor}{\textit{Out-of-domain (OOD)}}} \\
MMLU-Pro & 38.42 & \textbf{43.28} & 38.05 & \textbf{42.03} & 36.09 & \textbf{39.30} \\
GPQA-D & 25.19 & \textbf{26.07} & 23.36 & \textbf{27.27} & 23.74 & \textbf{23.93} \\
LCB v2 & 24.92 & \textbf{27.40} & 26.22 & \textbf{28.70} & 25.51 & \textbf{29.09} \\
IFEval & 44.36 & \textbf{46.04} & 42.57 & \textbf{44.60} & 38.49 & \textbf{41.97} \\
AlpacaEval & -0.26 & \textbf{-0.25} & -0.14 & \textbf{-0.10} & -0.02 & \textbf{0.02} \\
HaluEval & 46.60 & \textbf{48.76} & 46.75 & \textbf{47.85} & 46.72 & \textbf{47.42} \\

\textbf{Avg. ($\Delta$ vs. SFT)} & - & \textcolor{poscolor}{\textbf{+2.01}} & - & \textcolor{poscolor}{\textbf{+2.26}} & - & \textcolor{poscolor}{\textbf{+1.87}} \\
\bottomrule
\end{tabular}
\end{table*}

\begin{table*}[t]
\centering
\caption{Performance of \METHODNAME{} across SFT learning rates for Qwen2.5-7B fine-tuned on NuminaMath-CoT. The batch size is fixed at 256, and training lasts one epoch. Projection uses $\rho=0.5$, $\alpha=1$, and all weight matrices. Bold marks the best score in each pair. $^{\dagger}$ denotes the hyperparameters we used in the main experiment.}
\label{tab:qwen25_numina_learning_rate}
\scriptsize

\begin{tabular}{l*{3}{c>{\columncolor{oursbg}}c}}
\toprule
\multirow{2}{*}{\textbf{Benchmark}} & \multicolumn{2}{c}{\textbf{LR 2e-5}} & \multicolumn{2}{c}{\textbf{LR 5e-5}$^{\dagger}$} & \multicolumn{2}{c}{\textbf{LR 1e-4}} \\
\cmidrule(lr){2-3} \cmidrule(lr){4-5} \cmidrule(lr){6-7}
 & \textbf{SFT} & \textbf{\METHODNAME{}} & \textbf{SFT} & \textbf{\METHODNAME{}} & \textbf{SFT} & \textbf{\METHODNAME{}} \\
\midrule
\multicolumn{7}{l}{\textcolor{sectioncolor}{\textit{In-domain (ID)}}} \\
MATH500 & \textbf{45.60} & 44.80 & 44.84 & \textbf{49.36} & 39.41 & \textbf{49.21} \\
AIME25 & \textbf{0.42} & 0.00 & 0.83 & \textbf{1.46} & 0.63 & \textbf{1.25} \\
AIME24 & \textbf{1.67} & 1.46 & 1.04 & \textbf{1.46} & 1.04 & \textbf{1.25} \\
OlympiadBench & \textbf{14.83} & 14.29 & 14.08 & \textbf{17.72} & 10.98 & \textbf{18.18} \\
Minerva Math & \textbf{12.98} & 10.27 & 12.68 & \textbf{15.26} & 8.39 & \textbf{14.89} \\
AMC23 & 17.50 & \textbf{22.50} & 19.53 & \textbf{24.84} & 16.09 & \textbf{21.25} \\

\textbf{Avg. ($\Delta$ vs. SFT)} & - & \textcolor{poscolor}{\textbf{+0.05}} & - & \textcolor{poscolor}{\textbf{+2.85}} & - & \textcolor{poscolor}{\textbf{+4.92}} \\
\midrule
\multicolumn{7}{l}{\textcolor{sectioncolor}{\textit{Out-of-domain (OOD)}}} \\
MMLU-Pro & 35.26 & \textbf{40.09} & 38.05 & \textbf{42.03} & 38.18 & \textbf{42.14} \\
GPQA-D & 25.57 & \textbf{28.16} & 23.36 & \textbf{27.27} & 24.18 & \textbf{25.32} \\
LCB v2 & 29.22 & \textbf{30.53} & 26.22 & \textbf{28.70} & 12.46 & \textbf{19.96} \\
IFEval & 48.08 & \textbf{48.92} & 42.57 & \textbf{44.60} & 33.33 & \textbf{38.13} \\
AlpacaEval & \textbf{0.10} & -0.12 & -0.14 & \textbf{-0.10} & -0.92 & \textbf{-0.60} \\
HaluEval & 47.64 & \textbf{48.66} & 46.75 & \textbf{47.85} & 48.92 & \textbf{49.38} \\

\textbf{Avg. ($\Delta$ vs. SFT)} & - & \textcolor{poscolor}{\textbf{+1.73}} & - & \textcolor{poscolor}{\textbf{+2.26}} & - & \textcolor{poscolor}{\textbf{+3.03}} \\
\bottomrule
\end{tabular}
\end{table*}

\begin{table*}[t]
\centering
\caption{Performance of \METHODNAME{} across SFT training epochs for Qwen2.5-7B fine-tuned on NuminaMath-CoT. The learning rate is fixed at $5\times10^{-5}$, and the batch size is fixed at 256. Projection uses $\rho=0.5$, $\alpha=1$, and all weight matrices. Bold marks the best score in each pair.}
\label{tab:qwen25_numina_epochs}
\scriptsize

\begin{tabular}{l*{3}{c>{\columncolor{oursbg}}c}}
\toprule
\multirow{2}{*}{\textbf{Benchmark}} & \multicolumn{2}{c}{\textbf{Epochs 2}} & \multicolumn{2}{c}{\textbf{Epochs 4}} & \multicolumn{2}{c}{\textbf{Epochs 8}} \\
\cmidrule(lr){2-3} \cmidrule(lr){4-5} \cmidrule(lr){6-7}
 & \textbf{SFT} & \textbf{\METHODNAME{}} & \textbf{SFT} & \textbf{\METHODNAME{}} & \textbf{SFT} & \textbf{\METHODNAME{}} \\
\midrule
\multicolumn{7}{l}{\textcolor{sectioncolor}{\textit{In-domain (ID)}}} \\
MATH500 & 53.06 & \textbf{53.55} & \textbf{56.04} & \textbf{56.04} & 55.49 & \textbf{58.43} \\
AIME25 & \textbf{1.88} & 1.25 & \textbf{2.29} & 1.04 & 2.29 & \textbf{2.50} \\
AIME24 & 2.08 & \textbf{2.92} & \textbf{2.71} & \textbf{2.71} & \textbf{3.54} & 3.33 \\
OlympiadBench & 19.06 & \textbf{20.27} & 22.77 & \textbf{23.36} & 23.43 & \textbf{26.32} \\
Minerva Math & 16.38 & \textbf{17.03} & \textbf{19.65} & 18.73 & 19.19 & \textbf{20.68} \\
AMC23 & 26.41 & \textbf{28.28} & 28.12 & \textbf{30.31} & 28.44 & \textbf{33.59} \\

\textbf{Avg. ($\Delta$ vs. SFT)} & - & \textcolor{poscolor}{\textbf{+0.74}} & - & \textcolor{poscolor}{\textbf{+0.10}} & - & \textcolor{poscolor}{\textbf{+2.08}} \\
\midrule
\multicolumn{7}{l}{\textcolor{sectioncolor}{\textit{Out-of-domain (OOD)}}} \\
MMLU-Pro & 47.17 & \textbf{48.76} & 48.86 & \textbf{49.96} & 43.66 & \textbf{48.11} \\
GPQA-D & 25.09 & \textbf{28.03} & 28.85 & \textbf{30.84} & 26.36 & \textbf{30.93} \\
LCB v2 & \textbf{31.70} & 31.44 & \textbf{30.01} & 28.96 & 21.00 & \textbf{27.59} \\
IFEval & 46.04 & \textbf{47.24} & 44.84 & \textbf{47.60} & \textbf{47.48} & 47.36 \\
AlpacaEval & 0.15 & \textbf{0.19} & \textbf{0.44} & 0.40 & 0.36 & \textbf{0.60} \\
HaluEval & 51.22 & \textbf{52.57} & \textbf{58.42} & 57.30 & 56.33 & \textbf{56.70} \\

\textbf{Avg. ($\Delta$ vs. SFT)} & - & \textcolor{poscolor}{\textbf{+1.14}} & - & \textcolor{poscolor}{\textbf{+0.61}} & - & \textcolor{poscolor}{\textbf{+2.68}} \\
\bottomrule
\end{tabular}
\end{table*}

\subsection{Training-Time vs. Post-Hoc Projection}
\label{app:training_time_vs_post_hoc_projection}

We further compare different ways of incorporating the projection used by \METHODNAME{} into SFT. In addition to the default post-hoc \METHODNAME{}, we consider a projection-based training loss that penalizes updates within the pre-SFT principal subspace, as well as an online variant that applies the \METHODNAME{} projection every $K$ optimization steps during training. We evaluate $K=1$ and $K=100$, and include RPSFT as a related training-time regularization baseline. RPSFT regularizes the update jointly within the leading left and right singular subspaces, whereas our projection loss only penalizes updates along the leading right singular subspace. All experiments use Qwen2.5-7B trained on NuminaMath-CoT with the same training and evaluation settings as the main experiments.

As shown in Table~\ref{tab:training_time_vs_post_hoc_projection}, the default post-hoc \METHODNAME{} achieves the largest average gains on both ID and OOD benchmarks. Training-time projection loss improves OOD performance but provides little average ID gain, while online projection yields only modest improvements for both values of $K$. These results suggest that allowing SFT to learn its update first and calibrating it afterward is more effective than constraining or repeatedly projecting the update during optimization, which may interfere with the learning process.

\begin{table*}[t]
\centering
\caption{
Comparison of training-time and post-hoc projection on Qwen2.5-7B.
``Proj. Loss'' introduces projection-based regularization during SFT with weight $\lambda=1$.
``Online Proj.'' applies the projection every $K$ optimization steps and at the end of training.
\METHODNAME{} denotes the default post-hoc projection.
``Avg. ($\Delta$ vs. SFT)'' denotes the average score change relative to SFT.
\textbf{Bold} denotes the best score excluding the Base model.
}
\label{tab:training_time_vs_post_hoc_projection}

\scriptsize

\begin{tabular}{
l
>{\color{gray}}c
c
>{\columncolor{baselinebg}}c
>{\columncolor{baselinebg}}c
>{\columncolor{baselinebg}}c
>{\columncolor{baselinebg}}c
>{\columncolor{oursbg}}c
}
\toprule

\textbf{Benchmark}
& \textbf{Base}
& \textbf{SFT}
& \shortstack{\textbf{RPSFT}\\($\lambda=1$)}
& \shortstack{\textbf{Proj. Loss}\\($\lambda=1$)}
& \shortstack{\textbf{Online Proj.}\\($K=1$)}
& \shortstack{\textbf{Online Proj.}\\($K=100$)}
& \shortstack{\textbf{\METHODNAME{}}\\(post-hoc)}
\\

\midrule

\multicolumn{8}{l}{\textcolor{sectioncolor}{\textit{In-domain (ID)}}}
\\

MATH500
& 28.79 & 44.84 & 45.98 & 45.66 & 46.08 & 45.54 & \textbf{49.36}
\\

AIME25
& 0.42 & 0.83 & 1.04 & 0.63 & 0.42 & 0.63 & \textbf{1.46}
\\

AIME24
& 1.65 & 1.04 & \textbf{1.46} & 1.04 & 0.83 & 1.25 & \textbf{1.46}
\\

OlympiadBench
& 7.69 & 14.08 & 14.40 & 13.89 & 14.54 & 14.73 & \textbf{17.72}
\\

Minerva Math
& 9.26 & 12.68 & 13.01 & 12.73 & 12.80 & 12.64 & \textbf{15.26}
\\

AMC23
& 15.65 & 19.53 & 21.72 & 19.38 & 19.69 & 20.78 & \textbf{24.84}
\\

\textbf{Avg. ($\Delta$ vs. SFT)}
& -
& -
& \textcolor{poscolor}{+0.77}
& \textcolor{poscolor}{+0.06}
& \textcolor{poscolor}{+0.23}
& \textcolor{poscolor}{+0.43}
& \textbf{\textcolor{poscolor}{+2.85}}
\\

\midrule

\multicolumn{8}{l}{\textcolor{sectioncolor}{\textit{Out-of-domain (OOD)}}}
\\

MMLU-Pro
& 40.33 & 38.05 & 34.67 & 37.50 & 37.59 & 37.83 & \textbf{42.03}
\\

GPQA-D
& 26.77 & 23.36 & 24.15 & 25.73 & 24.59 & 23.48 & \textbf{27.27}
\\

LCB v2
& 28.96 & 26.22 & 29.03 & \textbf{29.75} & 28.44 & 27.40 & 28.70
\\

IFEval
& 42.57 & 42.57 & 44.48 & \textbf{46.76} & 43.41 & 44.00 & 44.60
\\

AlpacaEval
& -1.13 & -0.14 & -0.03 & \textbf{0.05} & -0.14 & -0.05 & -0.10
\\

HaluEval
& 46.86 & 46.75 & 46.86 & 47.42 & 45.64 & 46.62 & \textbf{47.85}
\\

\textbf{Avg. ($\Delta$ vs. SFT)}
& -
& -
& \textcolor{poscolor}{+0.39}
& \textcolor{poscolor}{+1.73}
& \textcolor{poscolor}{+0.45}
& \textcolor{poscolor}{+0.41}
& \textbf{\textcolor{poscolor}{+2.26}}
\\

\bottomrule
\end{tabular}
\end{table*}